\documentclass[preprint,12pt,authoryear]{elsarticle}

\usepackage{amssymb}
\usepackage{url}
\usepackage{amsmath}

\usepackage[utf8]{inputenc}
\usepackage{graphicx}
\usepackage{subcaption}
\usepackage{comment}
\usepackage{multirow} 
\usepackage{wrapfig} 
\usepackage{xcolor}
\usepackage{filecontents}
\usepackage[super]{nth}
\usepackage{hyperref}
\usepackage{amsmath}
\usepackage{lscape}
\usepackage{pifont}
\usepackage{array}
\usepackage{textcomp}
\usepackage{threeparttable}
\journal{Medical Image Analysis}

\begin{document}

\begin{frontmatter}
\title{SurgRIPE challenge: Benchmark of Surgical Robot Instrument Pose Estimation}

\author[1]{Haozheng Xu\corref{cor1}}
\ead{haozheng.xu19@imperial.ac.uk}

\author[1]{Alistair Weld}
\author[1]{Chi Xu}
\author[1]{Alfie Roddan}
\author[1]{Jo\~ao Cartucho}
\author[2,3]{Mert Asim Karaoglu}
\author[2]{Alexander Ladikos}
\author[1]{Yangke Li}
\author[4]{Yiping Li}
\author[5]{Daiyun Shen}
\author[7]{Geonhee Lee}
\author[7]{Seyeon Park}
\author[7]{Jongho Shin}
\author[8]{Lucy Fothergill}
\author[9]{Dominic Jones}
\author[9]{Pietro Valdastri}
\author[8]{Duygu Sarikaya}
\author[1]{Stamatia Giannarou}

\cortext[cor1]{Corresponding author}

\address[1]{The Hamlyn Centre for Robotic Surgery, Imperial College London, United Kingdom}
\address[2]{ImFusion GmbH, Munich, Germany}
\address[3]{Technical University of Munich, Munich, Germany}
\address[4]{Eindhoven University of Technology, Eindhoven, Netherlands}
\address[5]{Department of Engineering Physics, Tsinghua University, Beijing, China}
\address[7]{Department of Transdisciplinary Medicine, Seoul National University Hospital, South Korea}
\address[8]{School of Computing, University of Leeds, United Kingdom}
\address[9]{STORMLab, University of Leeds, United Kingdom}

\begin{abstract}
Accurate instrument pose estimation is a crucial step towards the future of robotic surgery, enabling applications such as autonomous surgical task execution. Vision-based methods for surgical instrument pose estimation provide a practical approach to tool tracking, but they often require markers to be attached to the instruments. Recently, more research has focused on the development of markerless methods based on deep learning. However, acquiring realistic surgical data, with ground truth (GT) instrument poses, required for deep learning training, is challenging. 
To address the issues in surgical instrument pose estimation, we introduce the Surgical Robot Instrument Pose Estimation (SurgRIPE) challenge, hosted at the 26th International Conference on Medical Image Computing and Computer-Assisted Intervention (MICCAI) in 2023. The objectives of this challenge are: (1) to provide the surgical vision community with realistic surgical video data paired with ground truth instrument poses, and (2) to establish a benchmark for evaluating markerless pose estimation methods.
The challenge led to the development of several novel algorithms that showcased improved accuracy and robustness over existing methods. The performance evaluation study on the SurgRIPE dataset highlights the potential of these advanced algorithms to be integrated into robotic surgery systems, paving the way for more precise and autonomous surgical procedures. The SurgRIPE challenge has successfully established a new benchmark for the field, encouraging further research and development in surgical robot instrument pose estimation.
\end{abstract}


\end{frontmatter}

\section{Introduction} \label{introduction}
Robot-assisted Minimally Invasive Surgery (RAMIS) has evolved significantly in the last decade driven by advances in artificial intelligence (AI) and surgical robotics. Platforms like the da Vinci\textregistered system have revolutionised surgical procedures by providing enhanced instrument control and intraoperative visualisation, greatly improving surgical assistance. Accurate pose estimation of surgical instruments has become a crucial task in RAMIS, as it is essential for enabling applications such as autonomous surgical task execution \citep{autoext}, surgical skill assessment \citep{jigsaws}, and surgical workflow analysis \citep{workflow}.

Commercial external devices, such as depth cameras and electromagnetic trackers \citep{hardward_track}, can provide accurate instrument pose estimation. However, their applicability intraoperatively is limited due to space requirements and hardware setup constraints in the Operating Room. Another solution for surgical instrument pose estimation is the use of kinematic information \citep{9981141} from the integrated joint encoders of robotic platforms such as the da Vinci\textregistered system. Although it does not rely on extra hardware, this method requires additional hand-eye calibration and suffers from estimation errors due to the complexity of the cable-driven robotic system \citep{10255755}.

Marker-based vision methods for surgical instrument pose estimation use external markers to simplify the task \citep{cartucho2021cylmarker}. They are limited by the fact that they rely on the marker always being visible in the camera’s Field-of-View (FOV) and are sensitive to background variations such as light reflection and occlusion. Moreover, these markers do not directly reflect the pose of the instrument, requiring the calculation of complex geometrical transformations. Therefore, markerless methods offer a promising and practical approach to surgical instrument tracking without hardware modifications.
\begin{table}[b]
\begin{center}
\resizebox{\linewidth}{!}{
\begin{tabular}{|c|c|c|c|} 
 \hline
 Dataset  &  Labelled Frames& Type & Annotation Type \\
 \hline
SurgRIPE & 2841 & Real Endoscope & 3D Pose \& Segmentation Mask\\
\hline
ROBUST-MIS \citep{ROBUST-MIS} & 10040 & Real Endoscope & Segmentation Mask \\
\hline
SuPer \citep{super} & 2000 & Real Stereo Endoscope & Kinematics Info \& Segmentation Mask \\
\hline
EndiVisPose \citep{endovispose} & 1850 &  Real Endoscope & 2D keypoints \& Segmentation Mask \\
 \hline
 EndoVis17 RobSeg \citep{endovis17robseg}  & 3000 & Real Endoscope & Segmentation Mask\\
 \hline
  EndoVis18 RobSeg \citep{endovis18_robseg}  & 5700 & Real Stereo Endoscope & Segmentation Mask\\
  \hline
\end{tabular}}
\caption{Comparison of the surgical tool localization datasets.}
\label{table:ds_compare}
\end{center}
\end{table}
Object pose estimation has been well studied in the computer vision literature, with benchmarks like LineMOD \citep{linemod} and YCB-Video \citep{xiang2018posecnn} utilising RGBD sensors and ArUco markers for 6 Degrees of Freedom (DoF) pose estimation in non-medical scenes. These datasets have facilitated significant advancements in pose estimation methods for natural scene tasks. However, a comparable benchmark for 6DoF pose estimation for surgical tasks and environments is lacking. 
Existing medical datasets focus on the processing of 2D information as shown in Table~\ref{table:ds_compare}. These datasets neglect the 3D information that is required for the estimation of 6DoF pose. For example, EndoVis18 RobSeg \citep{endovis18_robseg}, EndoVis17 RobSeg \citep{endovis17robseg} and ROBUST-MIS \citep{ROBUST-MIS} provide datasets with 2D segmentation annotations. EndoVisPose \citep{endovispose} provides data with ground truth 2D instrument keypoints. SuPer \citep{super} provides ground truth kinematic information. However, in the latter case, the relevant 6DoF pose cannot be derived directly from the kinematic information.

State-of-the-art (SOTA) methods, such as \citep{peng2019pvnet, Wen2023FoundationPoseU6}, have been established for natural scene 6DoF object pose estimation tasks. However, due to the lack of surgical benchmarks and datasets, translating these methods to RAMIS is difficult due to the following factors which are unique to surgery:
\begin{itemize}
    \item \textbf{Partial object visibility}. The limited operating space in RAMIS means that the endoscopic camera remains very close to the surgical instruments, allowing only partial visibility within the camera's field of view. This partial visibility hinders the performance of some state-of-the-art pose estimation methods due to the common requirements for full object visibility.
    \item \textbf{Surgical scene variations and occlusions}. In RAMIS, surgical tools interact with soft tissue and organs, leading to potential occlusions of the tool tip (e.g. due to blood), making pose estimation unstable. In addition, variations in the surgical scene, such as lighting conditions and specular reflections, further affect the accuracy of pose estimation.
    \item \textbf{High precision requirement}. Pose estimation datasets often use RGBD cameras to generate ground truth data, resulting in errors of centimetre scale. However, given that the typical diameter of surgical tools is around 5 millimetres, the accuracy requirements in RAMIS are of millimetre scale.
\end{itemize}

\begin{figure}[h]
    \centering
    \includegraphics[width=\columnwidth]{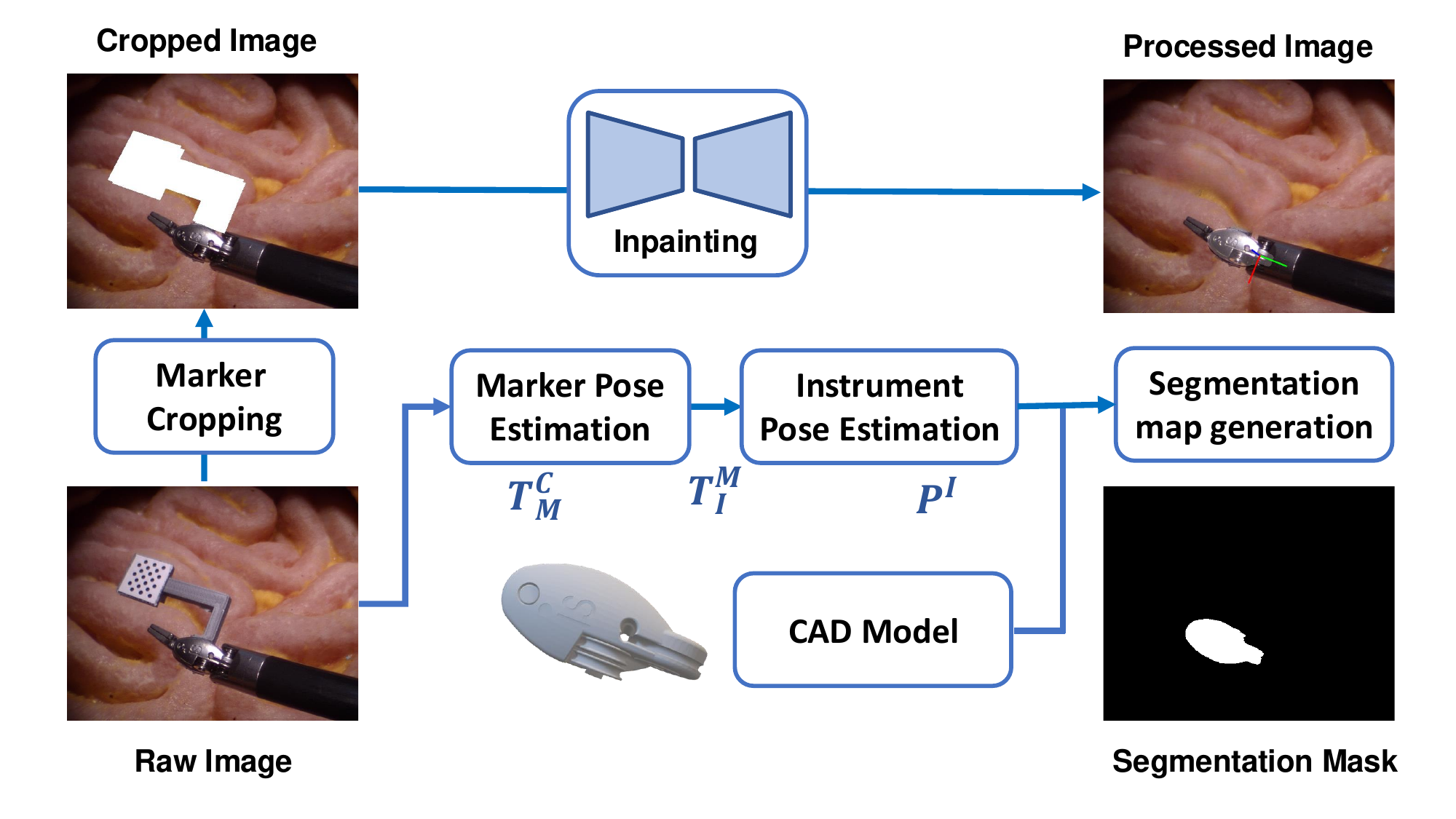}
    \caption{SurgRIPE data collection pipeline. The GT pose was captured with a keydot marker which was removed using image inpainting. The 3D model of the tool was used to generate the segmentation mask.
    }
    \label{fig:pipeline}
\end{figure}

To address the above issues, we present the SurgRIPE challenge, hosted at the 26th International Conference on Medical Image Computing and Computer-Assisted Intervention (MICCAI) in 2023. This paper first introduces the SurgRIPE dataset, which has been created for markerless estimation of the 6DoF pose of the wrist mechanism of surgical instruments. To acquire accurate and consistent ground truth surgical instrument poses while capturing video data in a realistic surgical setup, the novel pipeline shown in Fig.~\ref{fig:pipeline} was used which combines marker-based pose estimation with deep learning-based image inpainting. A keydot marker is used to get ground truth 6DoF pose data, which is then removed using a deep-learning inpainting model \citep{lama} to avoid generating any shortcut visual cues which could bias the pose estimation. Finally, 3D models are used to generate segmentation masks of the surgical instruments. 

\begin{figure}[h]
   \begin{minipage}{0.48\textwidth}
     \centering
     \includegraphics[width=\linewidth]{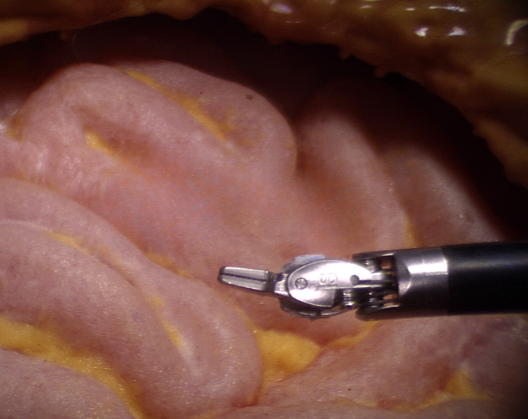}
     \caption{Non-occlusion Image Sample.}\label{fig:non_occ_sample}
   \end{minipage}\hfill
   \begin{minipage}{0.48\textwidth}
     \centering
     \includegraphics[width=\linewidth]{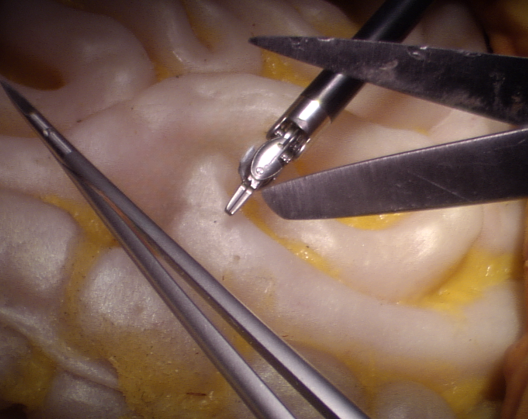}
     \caption{Occlusion Image Sample.}\label{fig:occ_sample}
   \end{minipage}
\end{figure}

The dataset includes video sequences to be used for two tasks, namely, pose estimation without occlusion and with occlusion as shown in Fig.~\ref{fig:non_occ_sample} and Fig.~\ref{fig:occ_sample}, respectively.
The six challenge participants proposed different markerless surgical instrument pose estimation methods which were validated on the SurgRIPE dataset.
The datasets, the benchmarking tool and the labeling tool are publicly available online and can be found at: \url{https://www.synapse.org/#!Synapse:syn51471789/wiki/}.

\section{Datasets and Annotation} \label{data}

\subsection{Data}

All SurgRIPE video data was captured using a da Vinci\texttrademark Si endoscopic stereo camera, ensuring the acquisition of high-quality and clinically relevant images. All the data was acquired in the Hamlyn Centre, Imperial College London. Only images from the left camera were preserved and used for processing. Two different da Vinci surgical instruments were used, namely, the Large Needle Driver (LND) and the Maryland Bipolar Forceps (MBF). The collected videos include sequences without and with instrument occlusions. The sequences without occlusion included variations in lighting conditions and background scenes to diversify the dataset and to simulate endoscopic environments such as when the light source is closer and farther from the tissue. The sequences with occlusion feature additional surgical tools, simulating scenarios with multiple instruments used by the surgeon.

The dataset is split into the LND and MBF subsets, with each subset containing only its respective instrument. Each subset contains the 3D model of the instrument, the pinhole camera intrinsic matrix, RGB images capturing the instrument moving over a surgical scene, segmentation masks of both the background and the instrument, and ground truth instrument poses. These poses correspond to the 6DoF movement of the wrist joint of the instruments.

The dataset information is illustrated in Table~\ref{table:dsinfo}. The LND dataset was extracted from 17 video clips containing 16747 raw frames and the MBF dataset was extracted from 15 video clips containing 11527 raw frames. All raw frames were captured at a framerate of 25 Hz. Finally, the LND subset consists of 1147 video frames without occlusion which are used for training, 373 frames without occlusion for testing, and 238 frames with occlusion for testing. The MBF subset contains 1,069 video frames without occlusion which are used for training, 209 frames without occlusion for testing, and 387 frames with occlusion for testing. In both cases, the occlusions were created using surgical instruments such as scissors and forceps.

\begin{table}[h]
\begin{center}
\resizebox{\linewidth}{!}{
\begin{tabular}{|c|c|c|c|c|} 
 \hline
 Instrument Type  & \text{             } LND (Large Needle Driver) \text{ 
            } & MBF (Maryland Bipolar Forcep )\\
 \hline
 Frame Rate & \multicolumn{2}{c|}{25} \\
 \hline
 Num of Raw Video Clips & 17 & 15 \\
  \hline
 Num of Raw Frames & 16747 & 11527 \\
  \hline
Resolution & \multicolumn{2}{c|}{$960 \times 540$} \\
 \hline
Training Frames & 1147 &  1069\\
 \hline
Test w/o Occlusion Frames & 373 & 209 \\
 \hline
Test w Occlusion Frames & 238 & 387 \\

 \hline
\end{tabular}}
\caption{Description of the generated dataset.}
\label{table:dsinfo}
\end{center}
\end{table}

To enable the participating teams to evaluate their algorithms before submitting the final results, sample images from the test sets were provided before the final submission deadline. The sample LND test sets include 5 frames without occlusion and 3 frames with occlusion. The sample MBF test sets also include 5 frames without occlusion and 3 frames with occlusion.

Both LND and MBF subsets contain the following data:

\begin{figure}[b]
    \centering
    \includegraphics[width=\columnwidth]{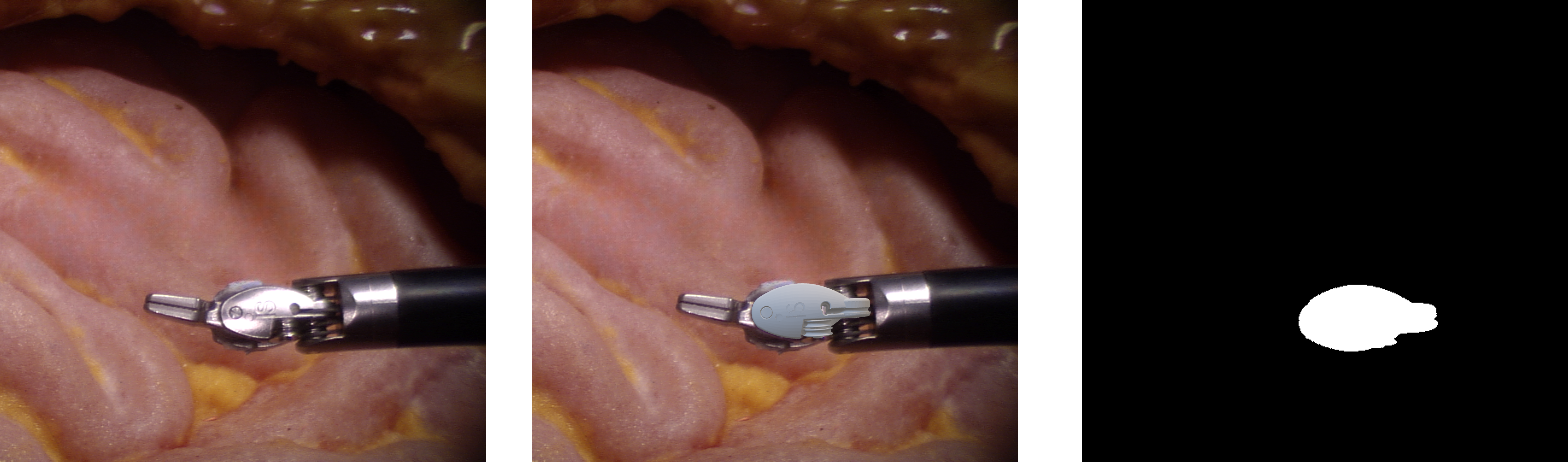}
    \caption{(Left) The raw RGB image. (Middle) Projection of the 3D model onto the raw image given the estimated 6DoF pose. (Right) The final segmentation result.}
    \label{fig:seg_generate}
\end{figure}

\begin{figure}[h]
    \centering
    \includegraphics[width=0.8\columnwidth]{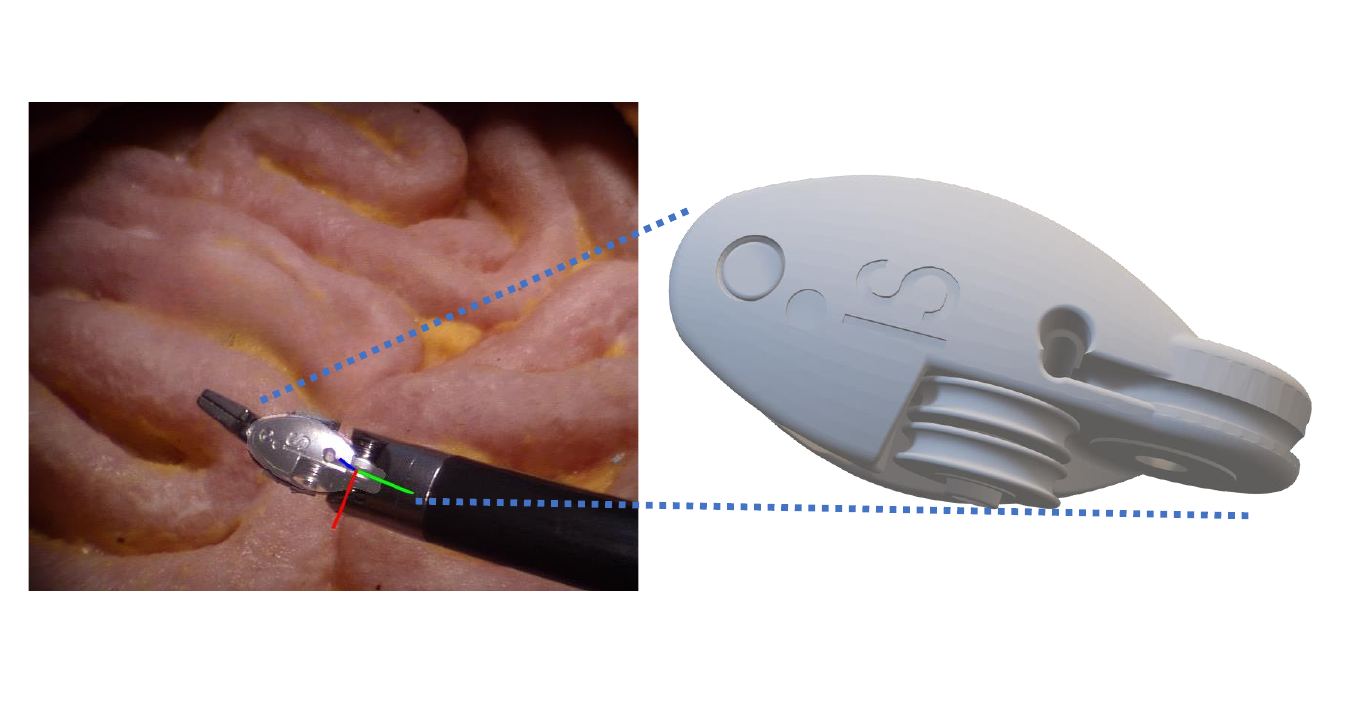}
    \caption{Projection of the 3D model onto the 2D image.
    }
    \label{fig:joint_3dmodel}
\end{figure}

\textbf{Instrument Model} 
The instrument 3D model is acquired from the da Vinci Research Kit (dVRK) wiki page \citep{6907809} which contains 3D models of multiple Endowrist instruments. This challenge focuses on the joint part in the 3D model, as shown in Fig.~\ref{fig:joint_3dmodel}.

\textbf{RGB Image}
The collected data includes frames with and without instrument occlusions. In the former category, the instrument is partially occluded due to the presence of different Endowrist\texttrademark instruments and surgical scissors, which have been used as occlusion objects, which is common in surgical scenarios. 
Fig.~\ref{fig:occ_sample} illustrates the occlusion caused by scissors in the presence of tweezers. As shown in Fig.~\ref{fig:non_occ_sample}, in frames without occlusion, the instrument is fully visible under different lighting conditions and backgrounds.

\textbf{Segmentation Mask}
The segmentation mask of the instrument is generated by projecting the 3D model of the instrument onto the 2D image given the camera's intrinsic parameters and the tool pose. In our case, only the joint part of each instrument is considered.
Fig.~\ref{fig:seg_generate} illustrates how the segmentation map is generated from the pose and 3D model.

\textbf{Instrument Pose}
The ground truth instrument pose consists of a rotation matrix (3x3) and a translation matrix (3x1). This pose corresponds to the 6DoF movement of the joint of the instrument, as shown in Fig.~\ref{fig:joint_3dmodel}. A special keydot pattern was used to obtain the ground truth pose of the surgical instruments. The pattern was then removed by applying inpainting to recreate a real surgical scenario where the keydot marker was not present.

\textbf{Camera Parameters}
We provide the camera intrinsic matrix as a 3x3 matrix. 
All the images have been undistorted, so the distortion parameters are set as None.

\begin{figure}[t]
    \centering
    \includegraphics[width=0.8\columnwidth]{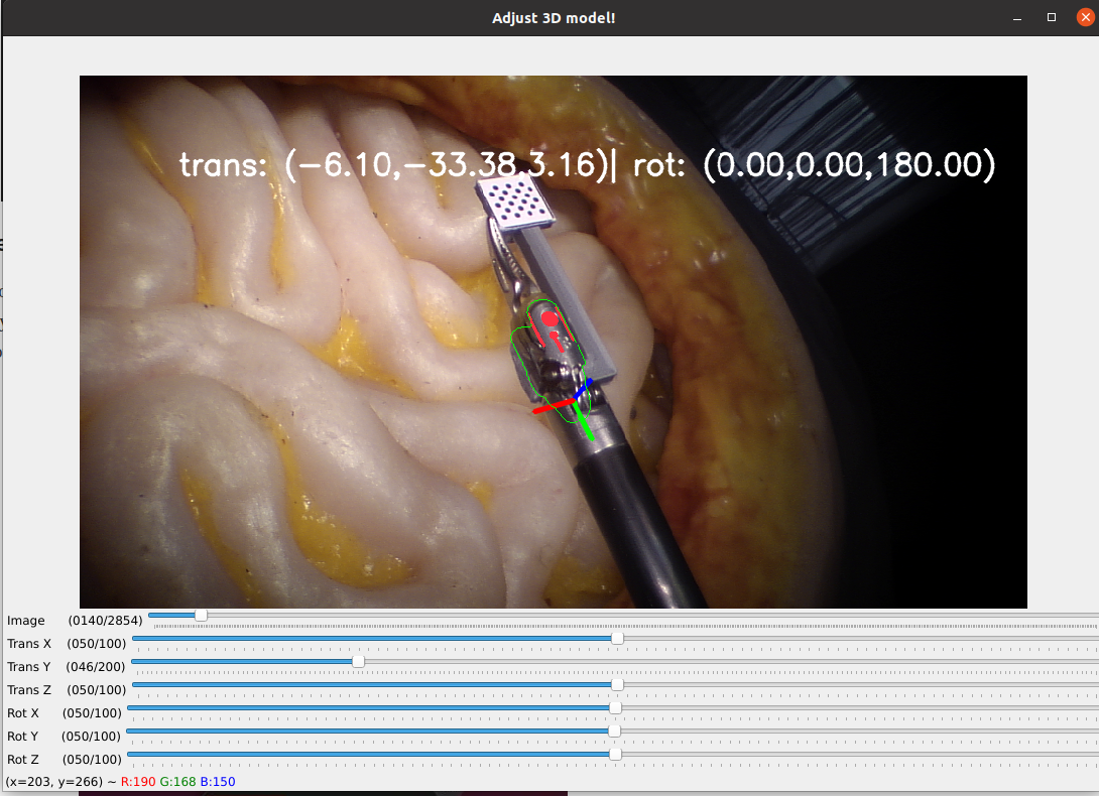}
    \caption{GUI for 3D Pose Alignment.}
    \label{fig:alignment}
\end{figure}

\subsection{Data annotation}\label{data annotation}
To estimate the ground truth of the 6DoF pose of a surgical instrument, we designed a holder with a keydot marker \citep{opencv_library} which is attached to the joint of the instrument as shown in the raw image in Fig.~\ref{fig:pipeline}. The pose of the keydot marker $T_M^C$ is estimated via a PnP solver \citep{epnp}. To recover the instrument pose $P^I$, the transformation from the keydot marker to the tip of the instrument $T_I^M$ needs to be estimated via the 3D pose alignment script as shown in Fig.~\ref{fig:alignment}.

Given the estimated marker pose and the camera intrinsic matrix, the 3D model of the instrument is projected onto the 2D image. Since the marker pose does not coincide with the instrument pose, the projection of the 3D model will not align with the target instrument area. To refine this 3D-to-2D projection, the Graphical User Interface (GUI) shown in Fig.~\ref{fig:alignment} has been implemented in OpenCV to manually adjust the transformation from the marker to the instrument tip $T_I^M$ until the instrument model projection perfectly aligns with the target area. The adjustment of the parameters of the marker-instrument transformation is repeated until the 3D model projection overlays the instrument on every frame of the dataset. This results in a precise transformation between the marker and the instrument, and therefore in accurate and consistent GT instrument pose estimation. Compared with some previous work \citep{8295119} that acquired manual annotation for every frame, our pipeline can achieve abundant high-precision and consistent ground truth annotations with minimal risk of human error.

\subsubsection{Image inpainting}
Since in a real surgical scenario, the keydot marker would not be present, we used image inpainting \citep{lama} to remove the keydot marker from the images used to train and test the compared models, as shown in Fig.~\ref{fig:pipeline}.
When inpainting was applied, we used random size masks to ensure that the method could not generate any consistent visual features that can be learnt by any computer vision method. To our knowledge, our proposed method is the first work that can generate a large surgical instrument 6DoF pose dataset of realistic images with high-accuracy ground truth annotations which can be used for training and testing of deep learning models.

\section{Benchmarking Metrics} \label{benchmarking}

The Python benchmarking toolkit developed as part of this challenge is open source and available online at: \url{https://github.com/CVRS-Hamlyn/SurgRIPETest}. 

\subsection{Performance evaluation protocol} 

Our benchmarking metrics include standard performance evaluation metrics (translation and rotation error) as well as the Benchmark for 6D Object Pose Estimation (BOP) metrics \citep{linemod}, which define a protocol to evaluate the pose estimation accuracy. BOP metrics provide comprehensive evaluation metrics and have been widely applied in pose estimation tasks. The Average Accuracy metric is also proposed here. We have split the performance evaluation metrics into primary and secondary. The former category is used to rank the compared methods.

\subsection{Primary Metrics} \label{Sec: PrimaryMetrics}

\textbf{ADD (Average 3D Distance)}
The Average 3D Distance of the model points (ADD) is defined as the mean 3D Euclidean distance between the ground truth and the predicted point cloud after applying the respective transformations. Let \(\mathbf{x}\) represent the set of 3D model points. The transformation is applied using both the ground truth pose, \((\mathbf{R}_{gt}, \mathbf{t}_{gt})\), and the predicted pose, \((\mathbf{R}_{pred}, \mathbf{t}_{pred})\), where \(\mathbf{R}\) denotes a rotation matrix and \(\mathbf{t}\) a translation vector.
Formally, the transformed model points using the ground truth and predicted poses are expressed as:
\begin{equation}
\mathbf{x}_{gt} = \mathbf{R}_{gt} \mathbf{x} + \mathbf{t}_{gt}, \quad \mathbf{x}_{pred} = \mathbf{R}_{pred} \mathbf{x} + \mathbf{t}_{pred}.
\end{equation}
The ADD metric is then computed as the mean Euclidean distance between the two transformed point clouds:
\begin{equation}
\text{ADD} = \frac{1}{m} \sum_{i=1}^{m} \| \mathbf{x}_{gt,i} - \mathbf{x}_{pred,i} \|_2,
\end{equation}
where \(m\) represents the number of points in the 3D model point cloud.
\newline
\textbf{Accuracy–ADD threshold Curve:} The Accuracy–ADD threshold Curve represents the accuracy for ADD thresholds varying from 0 to 10 mm.
This metric evaluates pose estimation performance across varying levels of precision by assessing how accurately the pose is estimated at different distance thresholds.
\newline
\textbf{Average Accuracy (Avg Acc) (0-5 mm):}
In BOP, the original accuracy of ADD is defined as the percentage of prediction error for a certain threshold equal to 10 \% of the model diameter of the object. Most surgical instruments have long cylindrical shapes with model diameters roughly equal to 40cm. Therefore, setting the ADD threshold to 10 \% of the surgical tool's model diameter would result in high error tolerance and provide a misleading representation of the model's performance. To deal with this issue prevalent in surgical applications, we define the average accuracy metric which is calculated by averaging the ADD errors for threshold values ranging from  0 mm to 5 mm.
\newline

\subsection{Secondary Metrics} \label{Sec: SecondaryMetrics}

\textbf{Translation \& Rotation Error:}
The 3D translation error (Euclidean mm) and rotation error (Euler degree) between the ground truth and the predicted poses are mathematically defined as:
\begin{equation}
    \text{Rotation Error} = \frac{1}{\sqrt{2}} \|\log(\mathbf{R}_{gt}^\top \mathbf{R}_{est})\|_F  
\end{equation}
\begin{equation}
\text{Translation Error} = \|\mathbf{t}_{gt} - \mathbf{t}_{est}\|
\end{equation}
\newline
\textbf{2D projection metric (proj2d):}
This metric evaluates how well the 3D points of an object match their ground truth 2D projections. The 3D points are projected onto the 2D image plane using the predicted pose and camera intrinsic parameters.
The aim is to measure the average distance between the corresponding projected points from the estimated pose and the ground truth pose.
Given the camera intrinsic matrix $\mathbf{K}$, the ground truth pose \((\mathbf{R}_{gt}, \mathbf{t}_{gt})\), and the predicted pose, \((\mathbf{R}_{pred}, \mathbf{t}_{pred})\), the point cloud $\mathbf{X}$ can be projected to the 2D point sets $\mathbf{p}^{\text{gt}}$ and $\mathbf{p}^{\text{pred}}$, respectively.
The proj2d error is computed as 
\begin{equation}
   \text{proj2D error} = \frac{1}{N} \sum_{i=1}^{N} \| \mathbf{p}_i^{\text{gt}} - \mathbf{p}_i^{\text{pred}} \|
\end{equation}
where, \(N\) is the total number of points in the point cloud.
The 2D projection metric is calculated as the percentage of frames in which the proj2d error is less than 5 pixels.
\newline
\textbf{5 mm 5-degree metric (mmd5):}
This metric is calculated as the percentage of frames where the Translation Error is below 5 mm and the Rotation Error is below 5 degrees.
\newline
\subsection{Winner identification protocol}
The teams were ranked according to their ADD and Avg Acc. The challenge winners are the submissions with the highest ADD and Avg Acc scores on the test dataset comprehensively. 

\section{Challenge organization}

All submissions were uploaded to the Synapse platform using Docker containers. Participants' code can be made available via emailing the corresponding team authors. The participants could submit multiple times before the submission deadline. 

All supervised and semi-supervised methods were allowed in this challenge. Considering there were abundant surgical instrument datasets proposed before, the participants were encouraged to use existing datasets along with SurgRIPE training data during the training phase.

Data presented in the challenge can be used for publication purposes only after the first version of the joint publication summarizing challenge results is submitted. All the team participants could be qualified as authors. The participants can publish their own method after the first version of the joint publication. The dataset is available to participating teams and other interested parties under CC BY 4.0 (Attribution).

The timetable for the challenge is shown in Table~\ref{table:timetable}. The training data (LND, MBF) was released on 2023 May 15th, followed by the sample test data on June 1st, allowing participants to familiarize themselves with the dataset and benchmark their solutions. The evaluation phase started on September 15th, during which participants were required to submit their solutions in the form of Docker containers. The final submission deadline was October 5th at 11:59 PM GMT. The challenge was hosted during the Endoscopic Vision Challenge workshop as part of MICCAI 2023 on October 8th, where results and insights were presented.

\begin{table}[h]
\begin{center}
\begin{tabular}{|c|c|} 
 \hline
Dates & Events\\
 \hline
2023 May, 15th &	Release of training data (LND, MBF)\\
 \hline
2023 June, 1st &	Release of sample test data\\
 \hline
2023 September, 15th	& Start of evaluation \\
 \hline
2023 October, 5th &	Submission deadline\\
 \hline
2023 October, 8th &	Challenge and Representation Day\\
 \hline
2024 December, 31th &	Acceptance of post-challenge submissions\\
 \hline
\end{tabular}
\caption{Timetable for the SurgRIPE challenge.}
\label{table:timetable}
\end{center}
\end{table}

The first-place award was £1000 and the second-place award was £500. The awards and rankings were announced at the EndoVis 2023 challenge workshop. Late submissions are included in this report, however, not eligible for awards. Members of the organizers' institutes could participate but were not eligible for awards.

The work presented in this paper follows the Biomedical Image Analysis Challenges (BIAS) Reporting Guideline \citep{BIAS} as required by MedIA.

\section{Challenge submissions} \label{submissions}

\subsection{ \textbf{[IGTUM]} ImFusion GmbH and Technical University of Munich, Munich, Germany}

The small size and the implicit symmetries existent within the targeted surgical tools make the task challenging. ImFusion's architecture decouples detection and pose estimation into two separate subtasks.

For object detection, You Only Look Once (YOLO) v5 \citep{yolov5} was used, which provides an effective solution for detecting 2D bounding boxes of object instances. Given the small number of training images, the training is based on a model pre-trained on the Common Objects in Context (COCO) dataset \citep{cocodataset}. More specifically, their medium-size model trained on higher resolution images, YOLOv5m6, was selected. For training, we extract the bounding boxes of the ground truth masks to provide labels. For validation, we randomly sample 20\% from the training set. Empirically decided, we train this model for 300 epochs with a batch size of 8. For data augmentation and other hyper-parameters, we follow the default, recommended settings of the original implementation of YOLOv5. 

For 6D object pose estimation, we employ SurfEmb \citep{9879714}. Unlike direct pose regression methods that require a high number of training samples capturing a large variation of poses of the object, SurfEmb uses surface coordinates of the object as the training target. This enables an efficient training scheme for the relatively low number of provided training samples. Furthermore, SurfEmb employs metric learning across a learned implicit representation of the object surface and their projections on the 2D images enabling an effective capture of the unknown and complex symmetries of the target objects. For training SurfEmb, the default settings proposed by the authors were followed. One single Residual Network (ResNet) 18 \citep{7780459} encoder was employed, for both surgical instruments (LND and MBF), coupled with separate decoders for each instrument. For the implicit representations of the object surfaces, we use a SIREN-based \citep{10.5555/3495724.3496350} multiple-layer perceptron (MLP) for each instrument. Given the low number of training samples, we separate only 20 samples for each instrument from the training set for validation. We train our model for 10 epochs with a batch size of 16 and employ the intensity and geometric data augmentations utilized in the original implementation. We use the default
values of the other hyper-parameters. 

Our approach is implemented using PyTorch and is trained on a single NVIDIA RTX 2080 GPU.

During inference, first YOLOv5 \citep{yolov5} detects an instance of the targeted surgical tool on the input image of size 960 x 540. This predicted bounding box is then used to crop a square patch around its center and resize it to 224 x 224. The convolutional neural network (CNN) part of our SurfEmb model takes in the cropped image along with the predicted class label and estimates 2D dense descriptors and the binary object mask. In addition, its MLP component predicts 3D descriptors on the densely sampled surface points of the targeted object 3D model. The 2D-3D matches are recovered through cosine similarity between the descriptors of the two domains. Since the surgical instruments have certain symmetries that can cause spatial ambiguities, we follow SurfEmb’s proposed multiple-hypothesis-based pose estimation strategy that creates numerous 2D-3D correspondence subsets and employs them within AP3P \citep{8099974} and select the maximum scoring pose considering the object segmentation mask, correspondence distribution and the probability of the visible surface coordinates. Finally, the predicted pose is refined through a render and compare strategy using Broyden–Fletcher–Goldfarb–Shanno (BFGS) optimizing for the maximization of the
2D-3D correspondence score.

\subsection{\textbf{[ICL]} Imperial College London, United Kingdom}
This method improves Pixel-wise Voting Network (PVNet) \citep{peng2019pvnet} by retaining its core pose estimation mechanism while introducing targeted data augmentation strategies and a de-glare algorithm.

To generate data augmentations, the authors simulated common occlusions in surgical environments by randomly generating holes in the training images, effectively enhancing the network's robustness in identifying and dealing with occlusions. This approach directly addresses the issue of visual obstruction caused by other tools or tissues in endoscopic surgery. To tackle the issue of image information loss or distortion caused by the reflection of surgical instruments, the authors applied a de-glare algorithm. This algorithm reduces glare effects, ensuring the integrity and continuity of image information, crucial for accurately identifying and locating surgical instruments with reflective surfaces. 

The improved method showed significant performance improvement in simulated endoscopic surgery settings, especially with obscured and reflective surfaces. However, when dealing with obstructions by other similar surgical tools, the algorithm's performance still fluctuated, highlighting the inherent challenges of pose estimation in surgical environments. 

Therefore, the author suggests that future work should explore modular approaches that handle mask segmentation and pose estimation separately, to ensure the accuracy of pose estimation is not affected by the accuracy of segmentation. This suggestion aims to provide a new direction for achieving more accurate and reliable pose estimation results in complex surgical scenarios. 

\subsection{\textbf{[TUDU]} Department of Engineering Physics, Tsinghua University, Beijing, China and, National University of Singapore, Singapore}
\begin{figure}[h]
    \centering
    \includegraphics[width=\columnwidth]{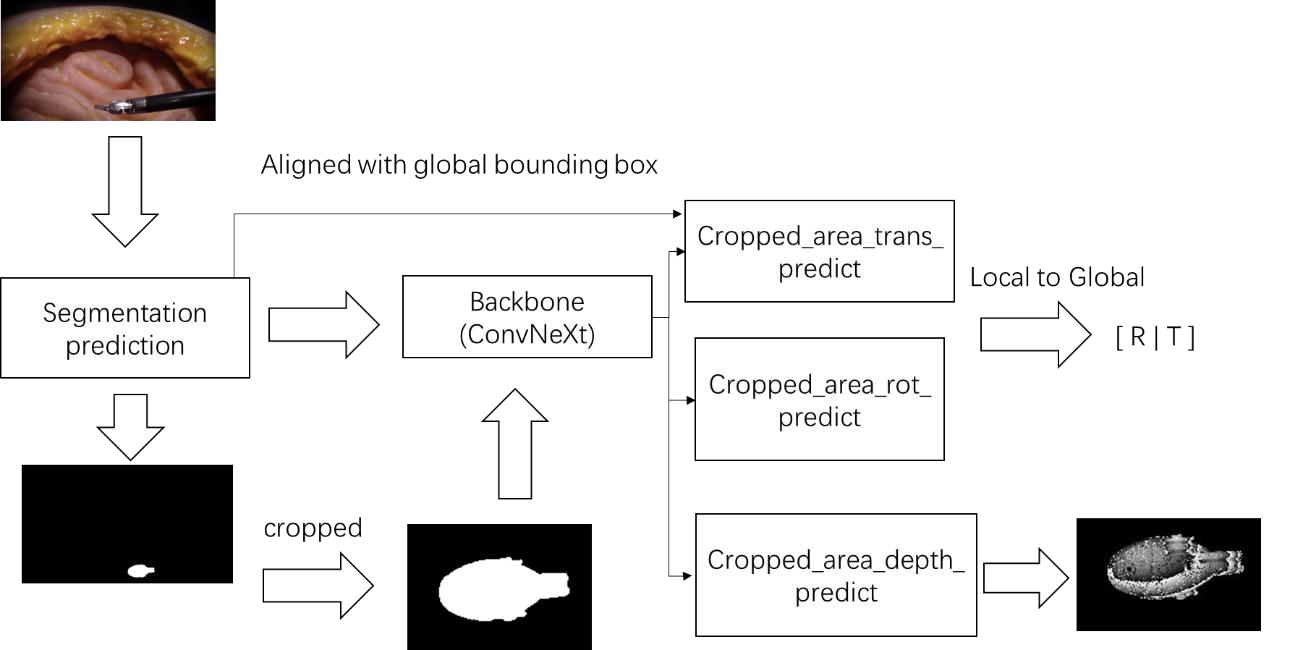}
    \caption{TUDU architecture diagram. The method includes (1) The segmentation branch which segments and crops the instrument. (2) The pose prediction branch where the backbone network predicts the tool pose based on the cropped regions.
    }
    \label{fig:daiyun_archi}
\end{figure}

Since the challenge provides 3D instrument models without RGB information, methods focusing on the geometric features of the instrument are prioritized. As the instrument size is relatively small compared to the background, a segmentation step is utilized to extract the object patch. However, cropping would cause the loss of global location information, so we adopted Scale Invariant Translation Estimation (SITE) to restore the global location.
The whole framework is illustrated in Fig.~\ref{fig:daiyun_archi}. As shown in the figure, this framework can be generally divided into (1) the segmentation step and (2) the pose prediction step. The bounding box of the instrument is derived from the segmentation result for further cropping and restoring global location. The depth map is predicted since such multi-tasking would enhance the performance of pose prediction.

We notice that the dataset contains images from diverse different scenes which would influence the prediction robustness without processing. So we adopt the copy-paste method to enlarge the dataset and reduce the influence of domain shift.

\subsubsection{The transformation of local and global coordinates}
The segmentation step produces masks with the 2D bounding box of an image region \([ \text{top}, \text{left}, \text{bottom}, \text{right} ]\). Our model's direct output is a \(\text{Batchsize} \times 7\) tensor. The first 4 elements correspond to a quaternion representation of the rotation. The last 3 elements \([ox, oy, oz]\) are camera coordinates. They can be transformed into the global translation \([x, y, z]\) according to the following relations:
\begin{align*}
x &= ox \times w + cx \\
y &= oy \times h + cy \\
z &= oz \times ratio
\end{align*}
where
\begin{align*}
w &= \text{right} - \text{left} \\
h &= \text{bottom} - \text{top} \\
cx &= (\text{right} + \text{left}) / 2 \\
cy &= (\text{top} + \text{bottom}) / 2 \\
ratio &= 960 \times 540 / (w \times h)
\end{align*}
\(960 \times 540\) is the original resolution of the image.

\subsubsection{Segmentation and Regression models}
DeepLabV3+ \citep{deeplabv3plus2018} is used for the segmentation step. We adopt ConvNeXT \citep{ConvNeXT} as the regression backbone. With the feature produced by the backbone, a pose vector and a depth map are predicted. This set of multi-tasking can enhance the pose estimation. The ground truth depth for the instrument region can be generated using the ground truth pose and the instrument model. The upper head is adopted as the head for depth map generation. The depth map is not used for refinement in post-processing step.

\subsubsection{Data pre-process}
As mentioned above, a copy-paste method is adopted. We use the segmentation mask to extract only instrument pixels and superimpose them onto images of different scenes. A background picture and a foreground picture are paired only if their bounding boxes do not overlap.

Besides copy-paste, brightness randomization is also used to further reduce the impact of domain shift.

\subsection{\textbf{[MVL\_3S]} Seoul National University Hospital, South Korea}

This challenge focuses on estimating the 6 Degrees of Freedom (6DoF) pose of a target object using RGB images exclusively. The distinctive aspect of this task lies in enabling the model to grasp the relationship between the camera and the object (Translation, Rotation). Drawing from our experience in a previous robot grasping challenge within the smart factory domain, we leveraged the problem-solving approach used back then to address this current challenge. For this task, we utilized the specialized EfficientPose architecture, tailor-made for estimating the 6DoF pose of a target object using RGB images only, without the need for depth information.

The EfficientPose \citep{bukschat2020efficientpose} architecture extends the capabilities of the EfficientNet model by incorporating two additional sub-networks for precise pose estimation. EfficientPose utilizes the EfficientNet as the backbone network and includes the Bi-directional Feature Pyramid Network (BiFPN) structure. This extension is tailored to perform efficient and accurate detection, achieving pose estimation performance using RGB images only.

EfficientPose integrates two distinct sub-networks, each designed for a specific task: 1. A sub-network dedicated to predicting coordinates and rotation, which is crucial for precise pose estimation. 2. A sub-network responsible for predicting the 2D bounding box of the object. By integrating these components, EfficientPose excels in providing accurate pose estimations.

\subsection{\textbf{[EUT]} Eindhoven University of Technology, the Netherlands}
This framework's network architecture follows an encoder-decoder structure, a design commonly employed for pose estimation tasks that consistently yields favourable outcomes. In the encoding phase, we adopt a modified version of ResNet-18 and incorporate a transformer encoder proposed by \citep{shaker2023swiftformer}. The Swift-Former, bolstered by its efficient additive attention mechanism, empowers the model to discern contextual relationships across distant image regions. It can enhance our model's ability to extract global context information by effectively integrating contextual cues from different regions. We believe this characteristic is valuable in this surgical tool pose estimation scenario, given the elongated and intricate nature of surgical tools. 

Similar to PVNet \citep{peng2019pvnet}, we train the network in a supervised manner to learn a semantic mask and a vertex map, which is a pixel-wise representation of the keypoints. The vertex map is stored as two channels: one for the $dx$ values and another for the $dy$ values across the entire image. Each keypoint has its corresponding vertex map with $dx$ and $dy$ values, resulting in a total of $2 \times k$ channels in the vertex branch of the network, with dimensions $H \times W$. Using the estimated semantic mask and unit vectors as inputs, we employ a RANSAC-based voting approach to generate potential keypoint hypotheses. After obtaining the estimated 2D keypoints that correspond to the sampled 3D keypoints, along with the camera intrinsics from the dataset, we use the \texttt{solvePnP} function in OpenCV to solve for the pose.
\subsection{Baseline for comparison}
\subsubsection{PVNet}
PVNet \citep{peng2019pvnet} is a state-of-the-art object pose estimation method which utilises pixel-wise unit vectors to estimate keypoints for keypoint detection. Then the detected keypoints can be used to solve a Perspective-n-Point (PnP) problem for pose estimation. A Pixel-wise Voting Network is introduced to localise pixel-wise unit vectors, then uses these vectors for keypoint voting using differentiable RANSAC method. This unit-vector representation is flexible for localising occlusion and truncated parts of objects, which is common in surgical scenes.

\section{Post-challenge submission}

\subsection{\textbf{[UOL]} University of Leeds, Leeds}
This approach used a pretrained ResNet-50 backbone for feature
extraction with additional fully connected layers for translation and rotation regression. The network
was trained using a multi-task loss function, which in addition to minimising rotation and translation
errors, aimed to align the transformed points of the 3D joint model and enforce consistency in the 3D
space.

\subsubsection{Methods}
\textbf{Preprocessing:}
RGB images were converted to tensor form and resized from (540, 960) to (224, 224) to fit with the ResNet image size requirements. The 3D translation vectors were converted from camera frame coordinates to image frame coordinates using the camera intrinsics and a 2D projection equation.
These coordinates were scaled; x and y coordinates were scaled by the size of the image 224 and z coordinates by 20. This was so that all values predicted by the model lay in the same range. To allow a more generalisable model on a test dataset with unseen images, some augmentations were also applied to the data, in the form of random 90-degree rotations, in both the clockwise and anti-clockwise directions, each with a 10 \% probability. As well as the images, these rotations were also applied to the ground truth rotation and translations.

\textbf{Implementation:}
We used a ResNet-50 encoder backbone for feature extraction, pretrained with the default ImageNet weights. We added a shared additional linear layer (in features = 1000, out features = 400), a linear
rotation layer (in features = 400, out features = 4) and a linear translation layer (in features = 400, out features = 3). We used a 90:10 train-validation random split for training.
A model was jointly trained on the Large Needle Driver (LND) and Maryland Bipolar Forceps (MBF)
surgical tool datasets, with a batch size of 8 and 4 for the training and validation sets, respectively. Each model was trained for 125 epochs, using an Adam optimiser and a learning rate of 0.0001.

\begin{figure}[h]
    \centering
    \begin{subfigure}{0.45\textwidth}
        \centering
        \includegraphics[width=\textwidth]{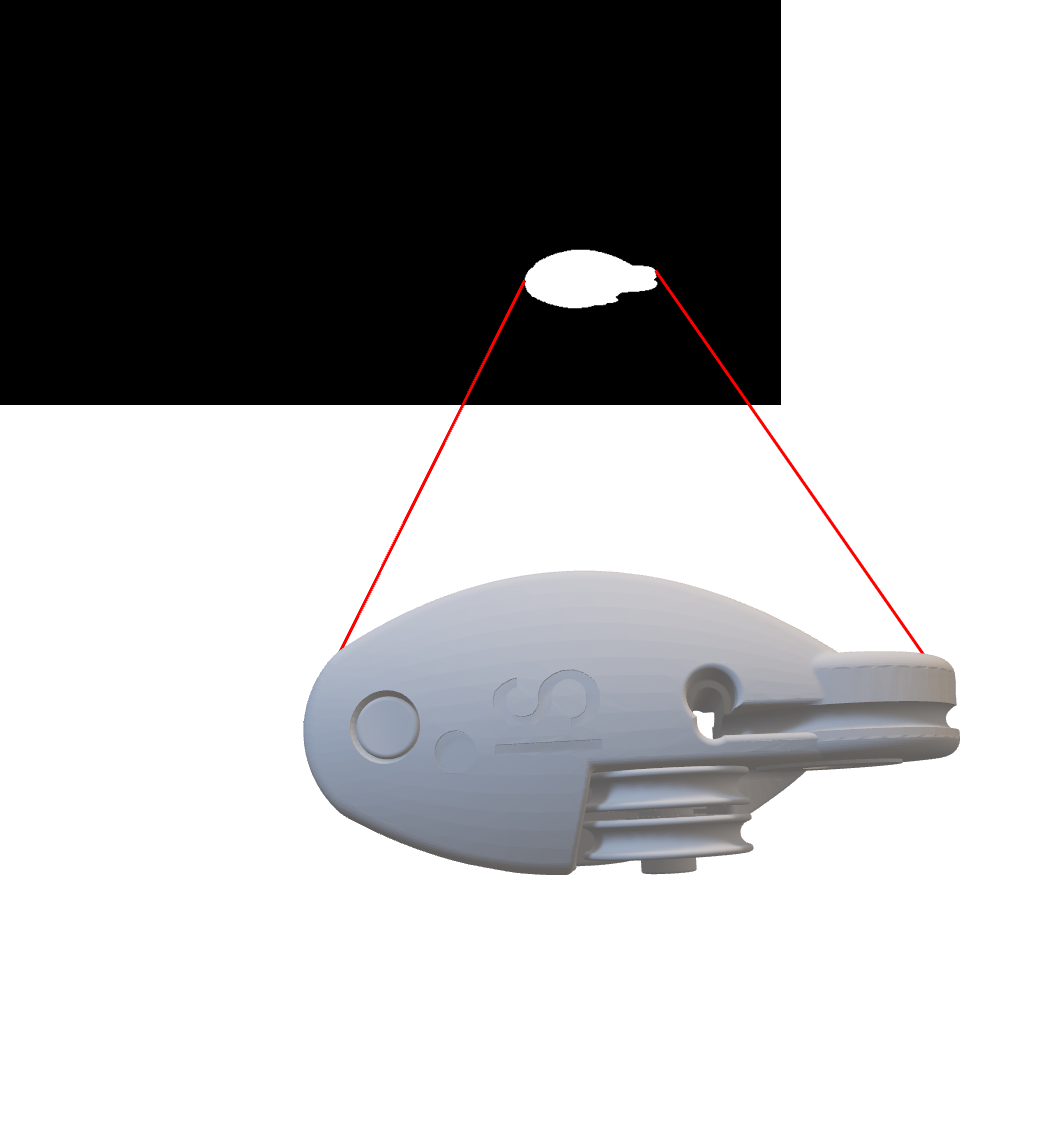}
        \caption{Visualisation of Projection Loss}\label{fig:proj_loss}
    \end{subfigure}
    \hfill
    \begin{subfigure}{0.45\textwidth}
        \centering
        \includegraphics[width=\textwidth]{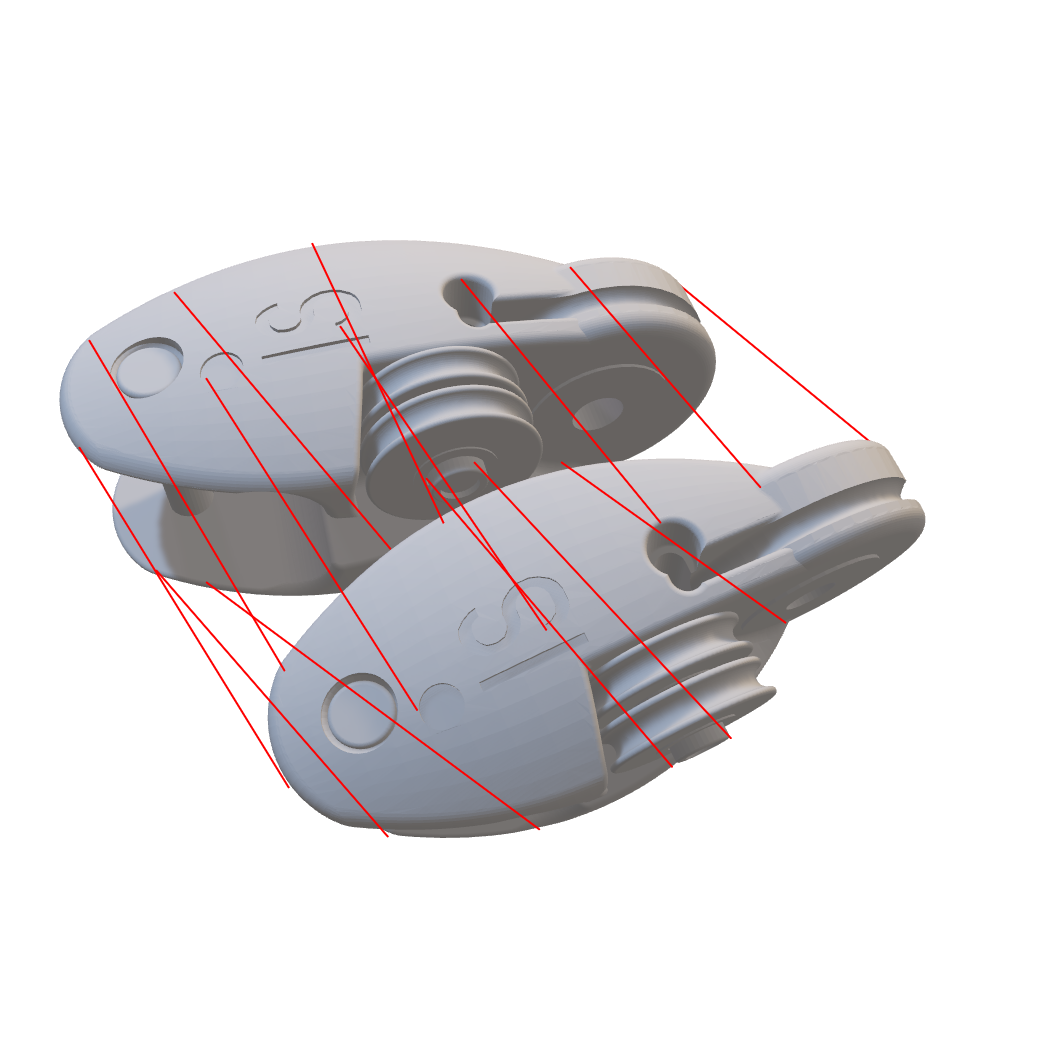}
        \caption{Visualisation of Point Loss}\label{fig:point_loss}
    \end{subfigure}
    \caption{Visualisation of Multi Losses in UOL Method.}
\end{figure}

\textbf{Loss Function:} We used a multi-task loss function that included points, projection, translation and contrastive loss terms. For the translation loss, the root mean square error (RMSE) was calculated between predicted and ground truth translation parameters and minimised. For the points loss, the 3D model of the tool joint was transformed by both the predicted and ground truth pose, and then the average L1 distance between corresponding points was calculated. This loss enforced consistency in 3D space and is shown in Fig.~\ref{fig:point_loss}. The projection loss term enforces consistency between the 3D and 2D spaces, as seen in Fig.~\ref{fig:proj_loss}. The 3D tool is transformed according to the ground truth and predicted poses, then projected into the 2D image plane using the camera intrinsics. The concave hull of the 2D points was then used to obtain a projected binary segmentation mask of the tool head and the dice loss was minimised between this and the ground truth segmentation mask.

The model performed better in both unoccluded tool datasets which is to be expected. In order to more robustly train the network to deal with occlusions, further augmentations to the data could be made, where parts of the image are at random covered with a mask. This would force the network to make more robust inferences. Furthermore, the model performed badly on images where 2 surgical tools appeared in the frame, specifically in the MBF TEST occ dataset. During training, the model seemed to perform much better when evaluated on the validation set compared to the test sets, specifically when
predicting rotation. This shows that the model failed to generalise beyond the training set. To address this in the future, additional augmentations could be performed on the training images to ensure the model is learning from a wider range of tool rotation depictions. Increasing the amount of training data by supplementing with synthetic images generated using tool renderings and artificial backgrounds could also improve the performance.

\section{Challenge results} \label{results}
There were 44 registered teams by the submission deadline. Five teams provided valid submissions before the deadline, and one team provided a valid submission after the deadline. All the submitted challenge results for each subset are presented in Tables \ref{table:subset1}-\ref{table:subset4}. We noticed that if the pose estimation fails in one frame, it will cause an outlier with a large translation and rotation error. This makes the average translation and rotation error of several methods heavily affected by these extreme error values. Therefore, to rank the competing methods, we mainly focus on the ADD, Accuracy–ADD threshold Curve and the Avg Acc as defined in Section \ref{Sec: PrimaryMetrics}. IGTUM was awarded first place, and ICL was awarded second place. In the results, late submissions have been denoted using an asterisk.

\begin{table*}[h]
\centering
\caption{Evaluation on LND Test Without Occlusion subset. Best 2 methods are in bold.}
\label{table:subset1}
\begin{threeparttable}
\begin{tabular}{|c|c|c|c|c|} 
 \hline
 Team  & ADD  & Avg Acc &Translation &Rotation\\
    & (10\% diameter) ↑ & (0-5 MM)↑ & Error (mm)↓ &Error (degree) ↓\\
 \hline
TUDU & 0.1314 &0.2336&6.3837&21.3342\\
\hline
IGTUM&\textbf{0.4182}&\textbf{0.5669}&\textbf{2.5618}&\textbf{5.1829}\\
\hline
ICL&0.1823&0.2657&63.3185&57.1676\\
\hline
EUT&0.1796&0.2665&44.5217&51.3492\\
\hline
MVL\_3S & 0.1156&0.2392&5.9092&27.2148\\
\hline
UOL*&0.0161&0.0640&8.1701&15.3407\\
\hline
PVNet**&\textbf{0.1930}&\textbf{0.2866}&46.7894&52.4488\\
 \hline
\end{tabular}
\begin{tablenotes}
    \item[$*$] post-challenge submission; \textit{Avg Acc:} Average Accuracy;
    \item[$**$] baseline method.
\end{tablenotes}
\end{threeparttable}
\end{table*}

\begin{table*}[h]
\centering
\caption{Evaluation on MBF Test Without Occlusion subset. Best 2 methods are in bold.}
\label{table:subset3}
\begin{threeparttable}
\begin{tabular}{|c|c|c|c|c|} 
 \hline
 Team  & ADD  & Avg Acc &Translation &Rotation\\
    & (10\% diameter) ↑ & (0-5 MM)↑ & Error (mm)↓ &Error (degree) ↓\\
 \hline
TUDU&0.1244&0.2145&5.8918&17.5108\\
 \hline
IGTUM&\textbf{0.3876}&\textbf{0.4441}&\textbf{3.0045}&\textbf{3.3593}\\
 \hline
ICL&\textbf{0.3684}&0.4223&5.3014&19.6242\\
 \hline
EUT&0.1483&0.1907&82.8653&65.6012\\
 \hline
MVL\_3S&0.3541&\textbf{0.4324}&\textbf{3.4532}&\textbf{10.3932}\\
\hline
UOL*&0.0239&0.0966&6.8555&10.4184\\
 \hline
PVNet**&0.3589&0.4198&3.5265&25.8501\\
 \hline
\end{tabular}
\begin{tablenotes}
    \item[$*$] post-challenge submission; \textit{Avg Acc:} Average Accuracy;
    \item[$**$] baseline method.
\end{tablenotes}
\end{threeparttable}
\end{table*}

\subsection{Instrument Pose Estimation Without Occlusion}
The instrument pose estimation performance without occlusion was evaluated across two test subsets namely, LND Test Without Occlusion and MBF Test Without Occlusion using the ADD (10\% diameter), Avg Acc (0-5 MM), Translation Error, and Rotation Error as shown in Table~\ref{table:subset1} and Table~\ref{table:subset3}. 

\begin{figure}[h]
    \centering
    \begin{subfigure}{0.45\textwidth}
        \centering
        \includegraphics[width=\linewidth]{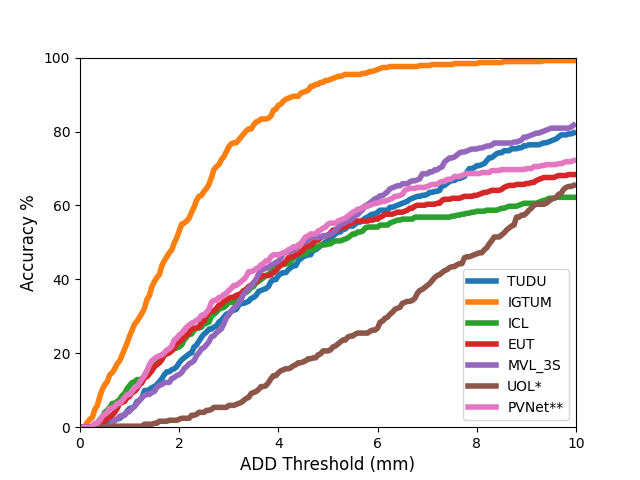}
        \caption{Curve on LND Test Without Occlusion}
        \label{fig:Acc_lnd}
    \end{subfigure}
    \hfill
    \begin{subfigure}{0.45\textwidth}
        \centering
        \includegraphics[width=\linewidth]{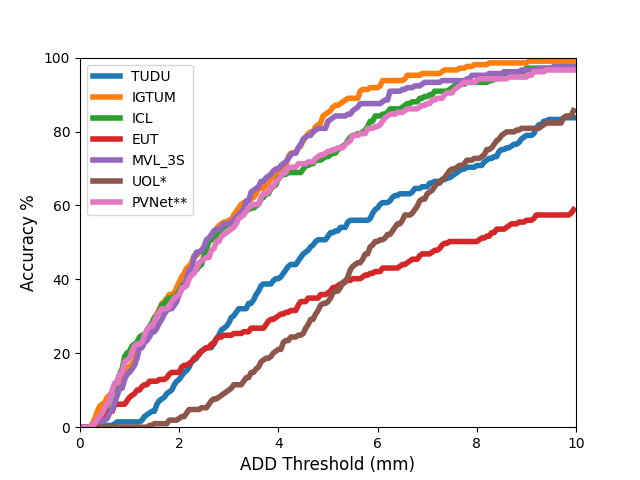}
        \caption{Curve on MBF Test Without Occlusion}
        \label{fig:Acc_mbf}
    \end{subfigure}
    \caption{Accuracy–ADD threshold Curves for test data in Testsets Without Occlusion.}
\end{figure}

Fig.~\ref{fig:Acc_lnd} and Fig.~\ref{fig:Acc_mbf} illustrate the Accuracy curve with respect to different ADD thresholds for the examined subsets. It can be noticed that the accuracy of IGTUM has a significant gap from the other compared methods for the subset LND Test Without Occlusion while IGTUM, ICL, MVL\_3S and PVNet have comparable accuracy for the MBF Test Without Occlusion subset.

\subsection{Instrument Tracking with Occlusion}
\begin{figure}[ht]
    \centering
    \begin{subfigure}{0.45\textwidth}
        \centering
        \includegraphics[width=\linewidth]{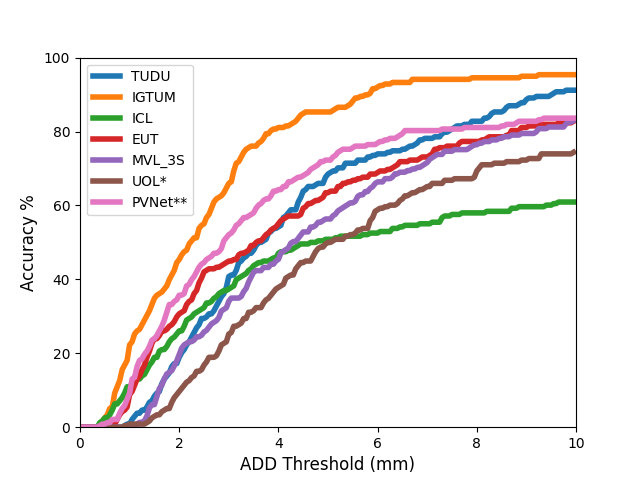}
        \caption{Curve on LND Test With Occlusion}
        \label{fig:Acc_lndocc}
    \end{subfigure}
    \hfill
    \begin{subfigure}{0.45\textwidth}
        \centering
        \includegraphics[width=\linewidth]{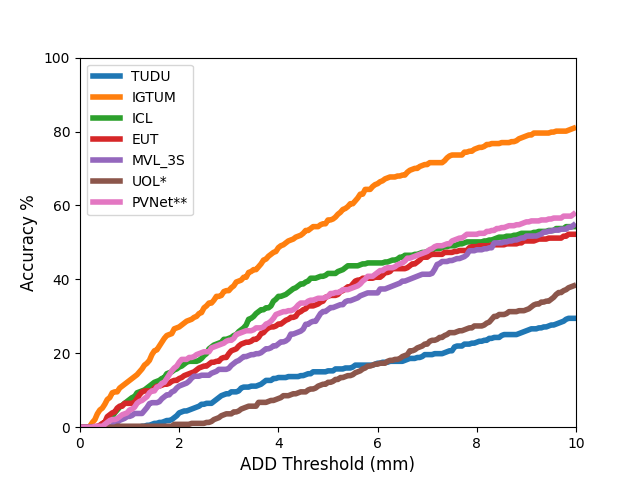}
        \caption{Curve on MBF Test With Occlusion}
        \label{fig:Acc_mbfocc}
    \end{subfigure}
    \caption{Accuracy–ADD threshold Curves for test data in Testsets With Occlusion.}
\end{figure}

Considering that pose estimation with occlusion is quite challenging, the performance of all methods under occlusion shows a significant drop as shown in Tables~\ref{table:subset2} and Tables \ref{table:subset4}. Across the LND Test With Occlusion and MBF Test With Occlusion subsets, IGTUM consistently outperformed the other methods with the highest ADD and Avg Acc scores and the lowest translation and rotation errors. This confirms its robustness and reliability under challenging conditions, such as occlusions. ICL, EUT and PVNet also showed promising performance in terms of ADD and Avg Acc scores, but with higher translation and rotation errors than IGTUM, indicating their limited effectiveness in challenging scenarios. TUDU and MVL\_3S had lower ADD and Avg Acc scores across both subsets, coupled with moderate to high translation and rotation error rates, suggesting that these methods are less reliable under occlusion. The above performance trends are verified by the Accuracy-ADD threshold curves shown in Fig.~\ref{fig:Acc_lndocc} and Fig.~\ref{fig:Acc_mbfocc}.

Overall, for both cases with and without occlusion, IGTUM outperformed the other methods, consistently achieving the highest accuracy (ADD and Avg Acc) and the lowest errors (Translation and Rotation) across all four subsets, indicating its robustness and effectiveness. It achieved 36.06\% ADD on average and 45.48\% Avg Acc for all subsets. MVL\_3S demonstrated good performance in the MBF Test Without Occlusion subset (Table~\ref{table:subset3}), with competitive ADD and Avg Acc scores and overall lower translation and rotation errors. It achieved 17.19\% ADD on average and 26.38\% Avg Acc for all four subsets. TUDU exhibited moderate accuracy overall with 10.09\% ADD on average and 20.15\% Avg Acc for all subsets. UOL achieved 4.04\% ADD on average and 14.63\% Avg Acc for all four subsets. To be noticed, MVL\_3S and TUDU achieved low translation errors but higher rotation errors. ICL, EUT and PVNet demonstrated strong performance in terms of ADD and Avg Acc scores but higher errors in translation and rotation error. EUT achieved 17.94\% ADD on average and 24.06\% Avg Acc while ICL obtained 22.92\% ADD on average and 29.52\% Avg Acc for all subsets.

The performance of the compared methods has also been evaluated on the secondary validation metrics, namely proj2d in Table~\ref{table:proj2d_table} and mmd5 in Table~\ref{table:mmd5_table} on all four subsets. IGTUM still outperformed the other methods with the best proj2d and mmd5 scores. ICL, EUT and PVNet demonstrated the second-best performance in the group since they followed a similar pipeline. UOL, TUDU and MVL\_3S had lower proj2d and mmd5 scores across both subsets.

\begin{table*}[h]
\centering
\caption{Evaluation on LND Test With Occlusion. Best 2 methods are in bold.}
\label{table:subset2}
\begin{threeparttable}
\begin{tabular}{|c|c|c|c|c|} 
 \hline
 Team  & ADD  & Avg Acc &Translation &Rotation\\
    & (10\% diameter) ↑ & (0-5 MM)↑ & Error (mm)↓ &Error (degree) ↓\\
 \hline
TUDU&0.1092&0.2930&\textbf{5.8365}&21.4826\\
 \hline
IGTUM&\textbf{0.3655}&\textbf{0.5044}&\textbf{5.4047}&\textbf{10.7144}\\
 \hline
ICL&0.2059&0.2899&66.5271&28.4566\\
 \hline
EUT&0.2605&0.3379&91.9081&40.8184\\
 \hline
MVL\_3S&0.1092&0.2497&7.5748&24.5983\\
\hline
UOL*&0.0378&0.1894&8.3454&17.7254\\
 \hline
PVNet**&\textbf{0.2731}&\textbf{0.3901}&28.0907&\textbf{17.5543}\\
 \hline
\end{tabular}
\begin{tablenotes}
    \item[$*$] post-challenge submission; \textit{Avg Acc:} Average Accuracy;
    \item[$**$] baseline method.
\end{tablenotes}
\end{threeparttable}
\end{table*}

\begin{table*}[h]
\centering
\caption{Evaluation on MBF Test With Occlusion Best 2 methods. are in bold.}
\label{table:subset4}
\begin{threeparttable}
\begin{tabular}{|c|c|c|c|c|} 
 \hline
 Team  & ADD  & Avg Acc &Translation &Rotation\\
    & (10\% diameter) ↑ & (0-5 MM)↑ & Error (mm)↓ &Error (degree) ↓\\
 \hline
TUDU&0.0388&0.0650&21.5032&37.9335\\
 \hline
IGTUM&\textbf{0.2713}&\textbf{0.3039}&\textbf{12.4446}&\textbf{18.4362}\\
 \hline
ICL&0.1602&\textbf{0.2027}&80.0850&39.2832\\
 \hline
EUT&0.1292&0.1675&62.7601&59.8497\\
 \hline
MVL\_3S&0.1088&0.1341&17.5545&34.1299\\
\hline
UOL*&0.0078&0.0346&15.9003&21.9342\\
 \hline
PVNet**&\textbf{0.1731}&0.1826&44.6732&31.0920\\
 \hline
\end{tabular}
\begin{tablenotes}
    \item[$*$] post-challenge submission; \textit{Avg Acc:} Average Accuracy;
    \item[$**$] baseline method.
\end{tablenotes}
\end{threeparttable}
\end{table*}

\begin{table*}[h]
\centering
\caption{Evaluation on proj2d. Best 2 methods are in bold.}
\label{table:proj2d_table}
\begin{threeparttable}
\begin{tabular}{|c|c|c|c|c|} 
 \hline
 Team  & LND Test  & MBF Test & LND Test & MBF Test\\
    & w/o Occlusion & w/o Occlusion & w Occlusion & w Occlusion\\
 \hline
TUDU&0.1126&0.0861&0.0252&0.0026\\
 \hline
IGTUM&\textbf{0.6944}&\textbf{0.9234}&\textbf{0.5294}&\textbf{0.7778}\\
 \hline
ICL&0.4718&\textbf{0.8086}&\textbf{0.4202}&\textbf{0.5814}\\
 \hline
EUT&0.3753&0.7847&0.1596&0.5065\\
 \hline
MVL\_3S&0.1075&0.4067&0.0175&0.1698\\
 \hline
 UOL*&0.0080&0.0042&0.0143&0.0129\\
 \hline
PVNet**&\textbf{0.5308}&0.8182&0.3487&0.5607\\
 \hline
\end{tabular}
\begin{tablenotes}
    \item[$*$] post-challenge submission;
    \item[$**$] baseline method.
\end{tablenotes}
\end{threeparttable}
\end{table*}

\begin{table*}[h]
\centering
\caption{Evaluation on mmd5. Best 2 methods are in bold.}
\label{table:mmd5_table}
\begin{threeparttable}
\begin{tabular}{|c|c|c|c|c|} 
 \hline
 Team  & LND Test  & MBF Test & LND Test & MBF Test\\
    & w/o Occlusion & w/o Occlusion & w Occlusion & w Occlusion\\
 \hline
TUDU&0.0134 & 0.0 & 0.0191 & 0.0052 \\
 \hline
IGTUM&\textbf{0.6273} & \textbf{0.3235} & \textbf{0.7608} & \textbf{0.4729} \\
 \hline
ICL&\textbf{0.19303} & 0.0210 & \textbf{0.5598} & \textbf{0.2842} \\
 \hline
EUT&0.0884 & \textbf{0.0504} & 0.2297 & 0.2145 \\
 \hline
MVL\_3S&0.0108 & 0.0 & 0.0813& 0.0318\\
 \hline
 UOL*&0.0134&0.0126&0.0718&0.0103\\
 \hline
PVNet**&0.1903 & \textbf{0.0504} & 0.4928 & 0.1757\\
 \hline
\end{tabular}
\begin{tablenotes}
    \item[$*$] post-challenge submission;
    \item[$**$] baseline method.
\end{tablenotes}
\end{threeparttable}
\end{table*}

\section{Analysis of the results} \label{result_analysis}

\begin{figure}[htbp]
    \centering
    \includegraphics[width=\columnwidth]{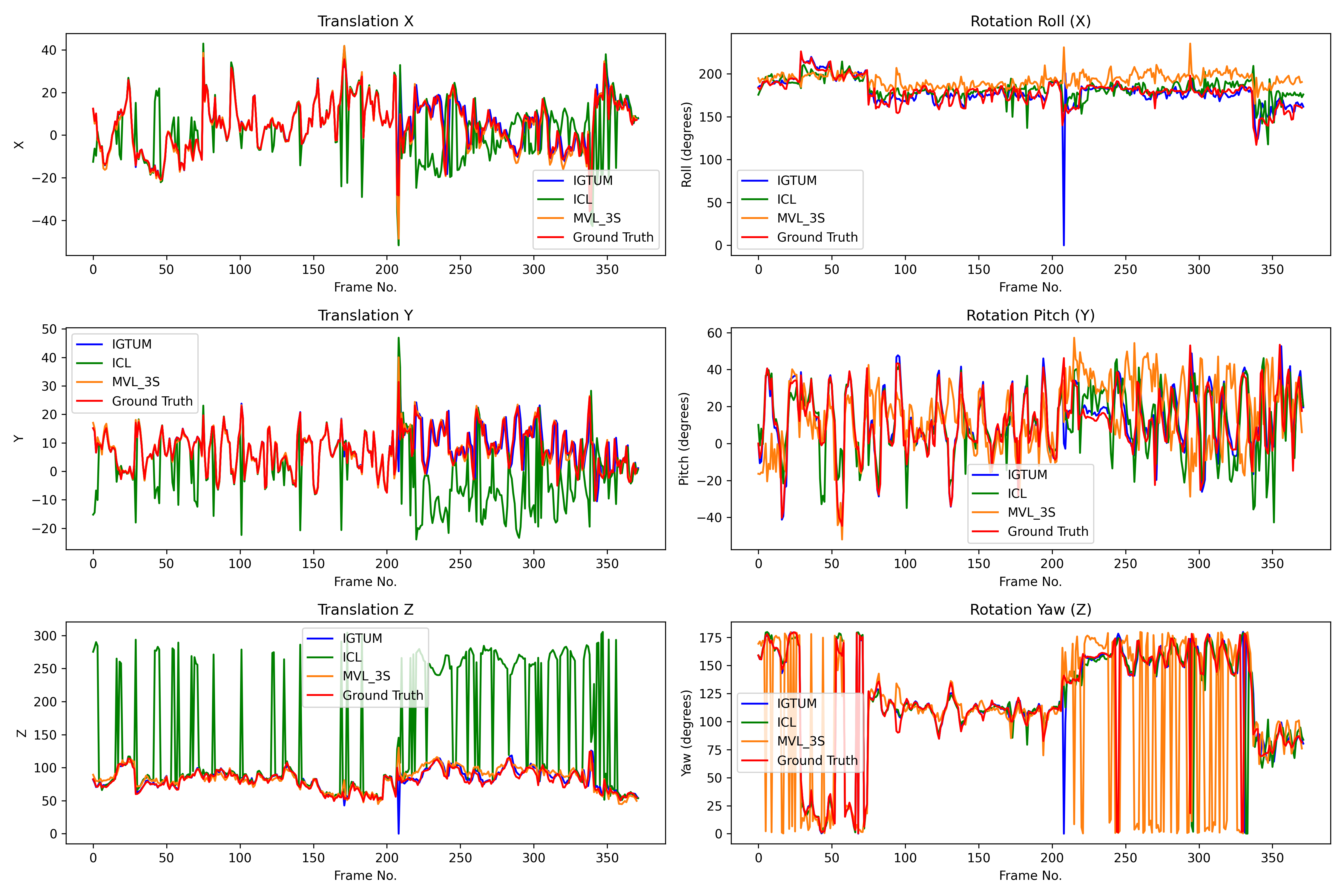}
    \caption{Trajectory comparison in LND Test Without Occlusion.}
    \label{fig:lnd_test_6dof}
\end{figure}

\begin{figure}[htbp]
    \centering
    \includegraphics[width=\columnwidth]{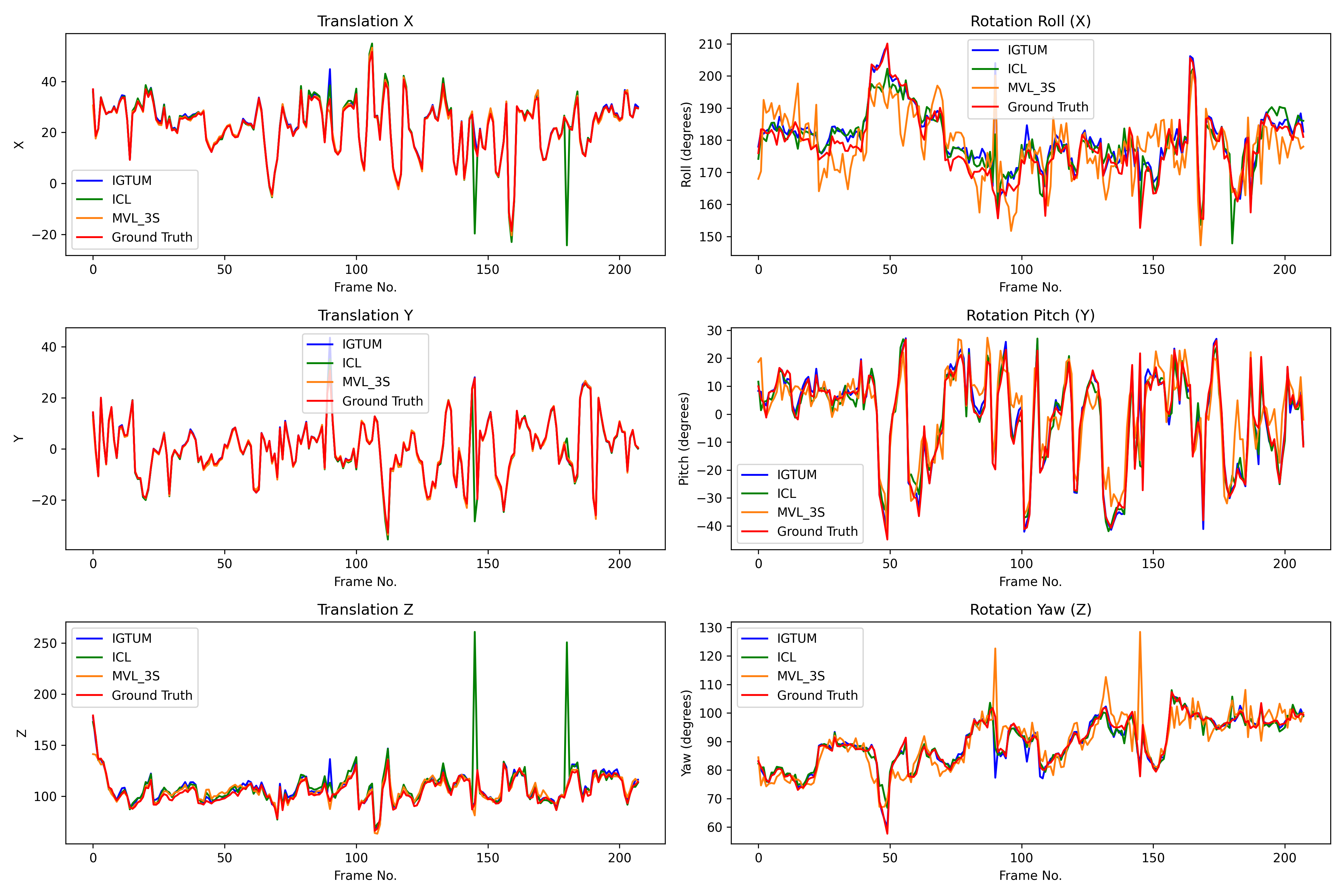}
    \caption{Trajectory comparison in MBF Test Without Occlusion.}
    \label{fig:mbf_test_6dof}
\end{figure}

\begin{figure}[htbp]
    \centering
    \includegraphics[width=\columnwidth]{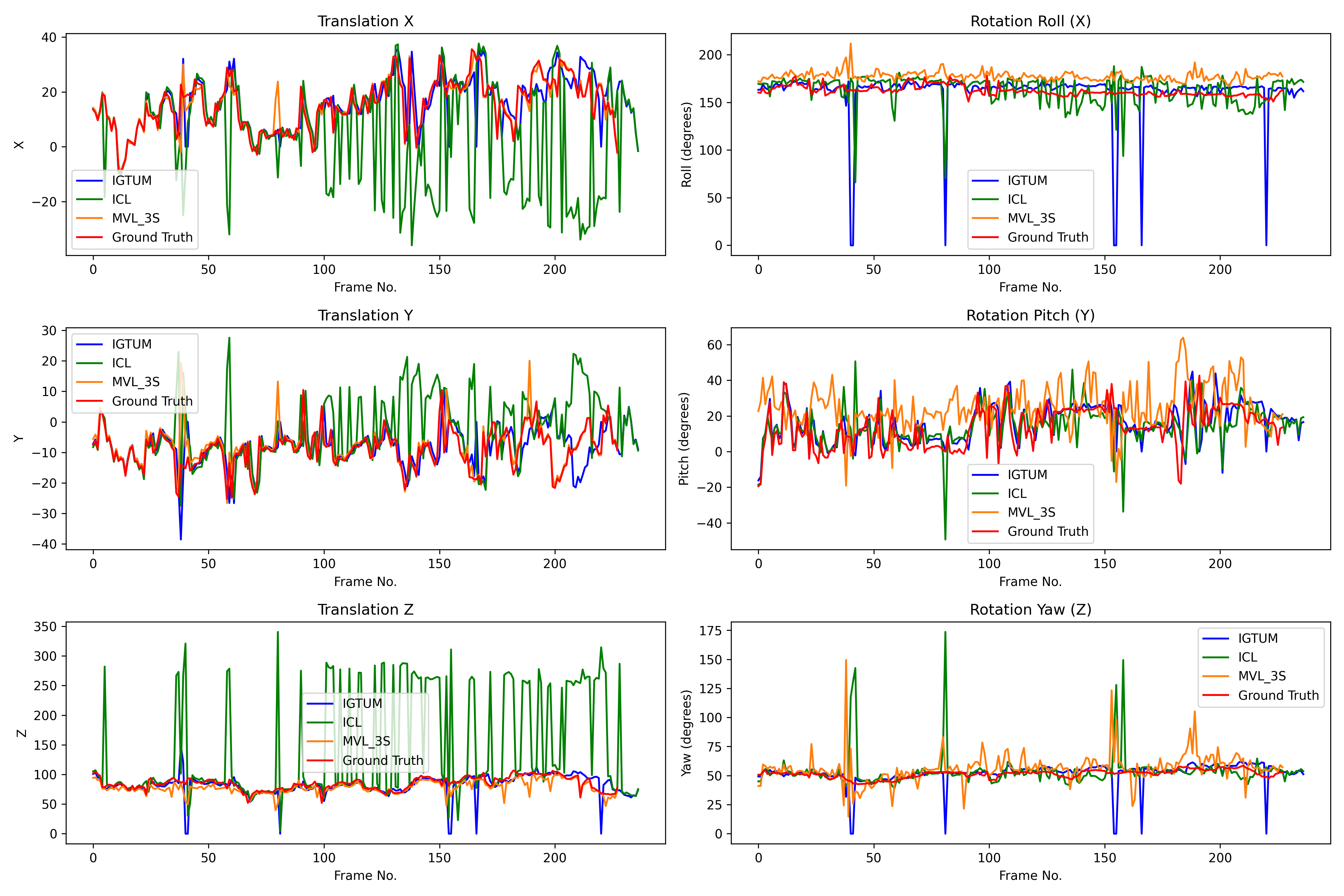}
    \caption{Trajectory comparison in LND Test With Occlusion.}
    \label{fig:lnd_test_occ_6dof}
\end{figure}

\begin{figure}[htbp]
    \centering
    \includegraphics[width=\columnwidth]{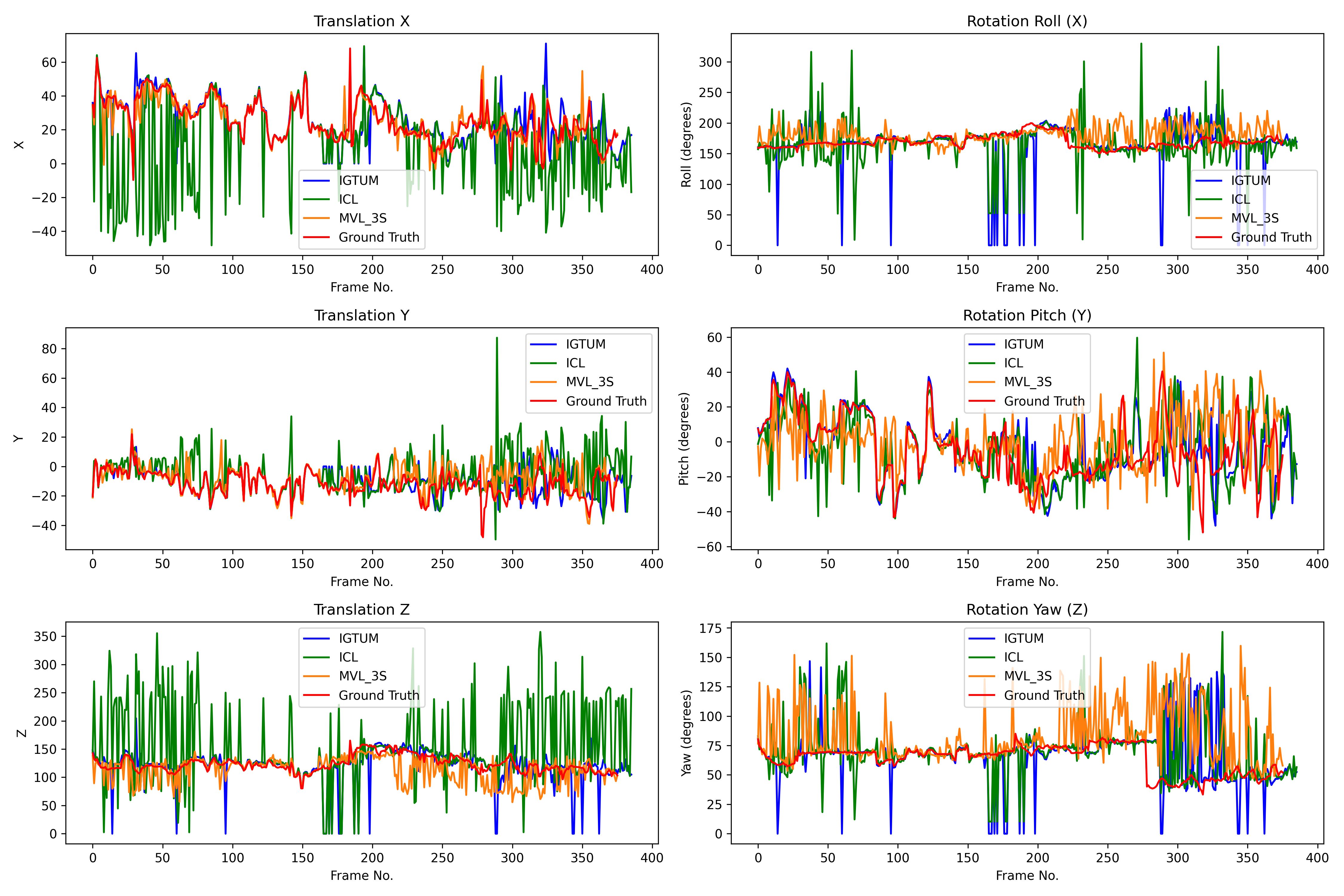}
    \caption{Trajectory comparison in MBF Test With Occlusion.}
    \label{fig:mbf_test_occ_6dof}
\end{figure}

The results of this challenge highlight the ability of deep learning techniques to deal with the task of surgical instrument pose estimation even under challenging scenarios such as the presence of occlusion. According to the architecture employed, these submitted methods can be broadly categorised into direct prediction methods (MVL\_3S, TUDU, UOL), two-stage methods with intermediate steps (ICL, EUT, PVNet), and the method that uses candidate hypotheses generation (IGTUM). To provide an intuitive understanding of the 6DoF poses estimated for a video sequence, we visualize the corresponding translations and rotations in the 3D space in Fig.~\ref{fig:lnd_test_6dof}-\ref{fig:mbf_test_occ_6dof}.
To ensure clarity and readability in visualizing the data, only three representative methods were plotted namely, IGTUM, ICL, MVL\_3S.

In the first category, TUDU and MVL\_3S aim to estimate the 3D translation and rotation of the target object directly from image features in a single forward pass. They use segmentation to generate a 2D bounding box for 2D localisation. The direct estimation pipeline offers simplicity but often struggles with ambiguities, complex rotations, and occlusions due to the absence of intermediate geometric reasoning. These limitations cause the higher rotation error and performance drop under occlusion, which is reflected by the low ADD and Avg Acc metrics demonstrated in results of the previous section. As shown in the 6DoF trajectories in Fig.~\ref{fig:lnd_test_6dof}-\ref{fig:mbf_test_occ_6dof}, MVL\_3S achieved a larger average rotation error than IGTUM and ICL. 

Two-stage methods, like ICL, EUT and PVNet, involve extracting intermediate representations, such as keypoints or segmentation masks, before the final pose estimation. Instead of using a deep learning model to estimate the pose directly, these models use intermediate results and the PnP solver \citep{epnp} to compute 2D-3D keypoint correspondences which are processed further for 6DoF pose estimation. The two-stage architecture makes these methods more robust to occlusions and complex geometries. However, errors in the intermediate results like incorrect localisation of a single keypoint in the first stage, can propagate to the second stage in the PnP solver, resulting in a significant overall error. This property is also reflected in their high Average Accuracy and high translation and rotation error in the results section. According to the estimated trajectories shown for the Z-axis translation estimation in Fig.~\ref{fig:lnd_test_6dof}-\ref{fig:mbf_test_occ_6dof}, two-stage methods like ICL generate more frequent outliers instead of consistent trajectories compared to the direct prediction methods. 

The last category which includes the IGTUM method, uses candidate hypotheses. This method generates multiple possible poses for an object and refines them using scoring mechanisms or geometric consistency checks. For example, IGTUM utilises segmentation to localise the instrument area on the image plane, then uses a feature detector to extract visual features, and finally selects the most likely rotation-translation candidates from the potential tool pose space. These methods are highly accurate and robust, especially in handling ambiguities, occlusions, and cluttered environments. As a result, IGTUM outperformed other methods in the results section.



\section{Discussion} \label{discussion}
\subsection{Role of Vision-based Deep Learning in Surgical Instrument Pose Estimation}
The participants' results demonstrate the promising potential of applying advanced deep learning technology in surgical instrument pose estimation, especially in the occluded scenarios which are common but challenging in real surgery. Compared with the traditional tracking methods relying on extra hardware and markers, the vision-based methods provide more flexibility and cost-efficiency. However, there is still clear scope for improvement in further application. Although some two-stage methods could provide intermediate results, deep learning methods still lack interpretability and uncertainty estimation. In addition, all the participants' methods require retraining for every new instrument, while the ground truth 6DoF pose annotation is quite time-consuming and limited by the instrument size. In the future, generalizable pre-trained model could be implemented to solve the issue.

\subsection{Ground truth data accuracy and error analysis}
\begin{figure}[h]
    \centering
    \includegraphics[width=\columnwidth]{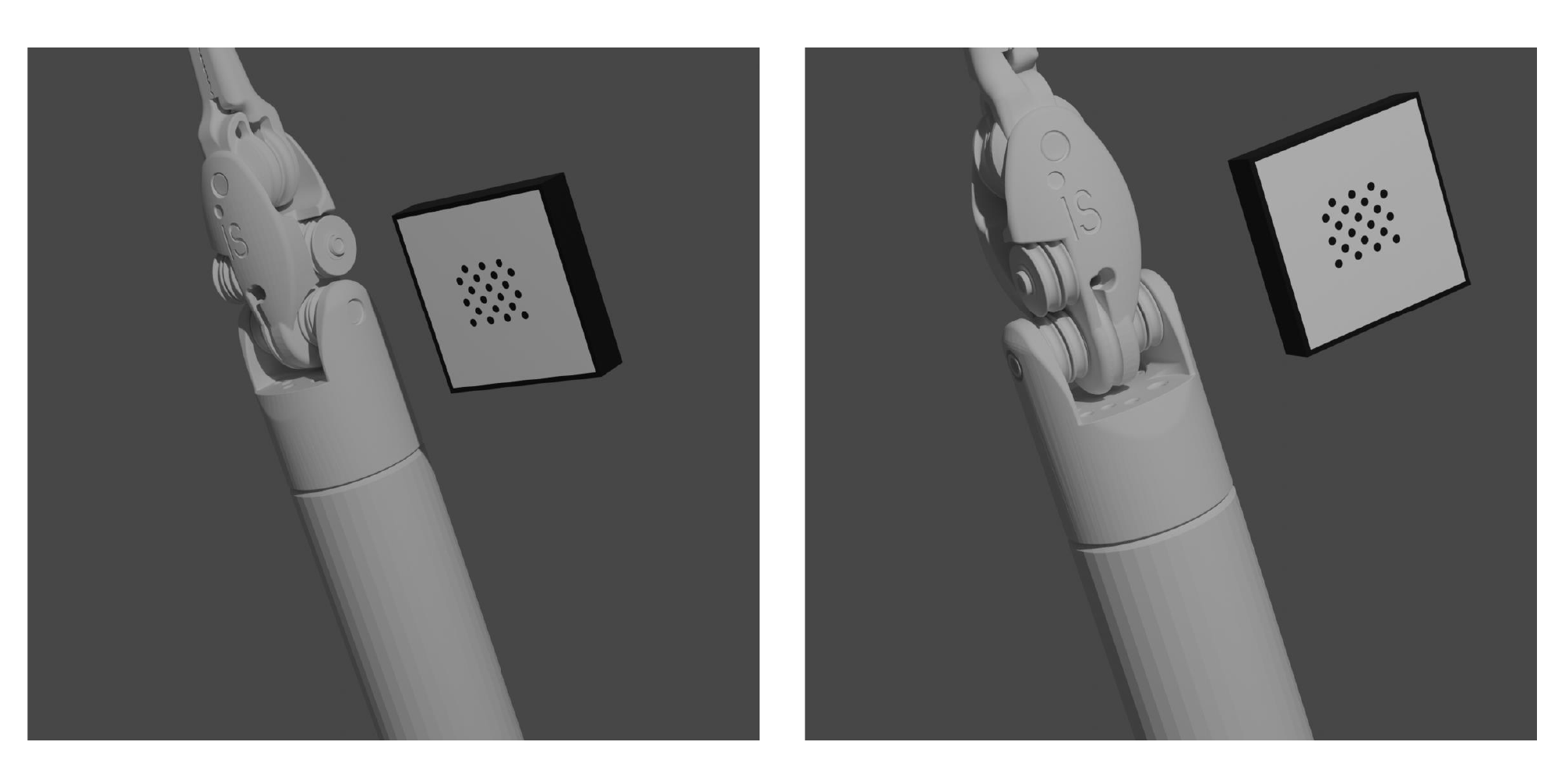}
    \caption{Simulation sample images.
    }
    \label{fig:simu_sample}
\end{figure}

To evaluate the accuracy of our ground truth pose generation, we created simulated images with known instrument pose information. More specifically, we used the VisionBlender simulation platform \citep{cartucho2020visionblender} to generate images of the instrument with a keydot marker, accompanied by their actual pose. As shown in Fig.~\ref{fig:simu_sample}, the keydot marker moved along with the tool 3D model. To guarantee consistency between the simulation images and real endoscopic images, the same camera intrinsic matrix was used as in the LND subset. To generate ground truth tool pose estimation for the simulated data, we estimated the pose of the keydot marker by analysing the simulation images and followed the procedure explained in Section \ref{data annotation}. Among the 50 samples of simulated data, the translation error between the estimated and actual instrument pose is 0.253 mm and the rotation error is 0.302 degrees. This verifies the accuracy of our generated ground truth data. 

\subsection{Background Scene Simulation}
The background scene was simulated with a realistic high-fidelity abdominal phantom in our experiments. To create diversity of the background conditions, video data was captured from different areas of the phantom corresponding to different organs, using the Da Vinci Si camera. Varying lighting conditions were generated by adjusting the light source intensity on the Da Vinci between 40\% and 100\% with a step of 10\%. In addition, a ceiling light source was used to illuminate our scene. We adjusted the intensity and the wavelength of the light source, and we also moved it to different locations above our experimental set up to further increase the variability of the lighting conditions in our captured video data.

\subsection{Monocular Image Capture}
In this work, we chose to capture only the left camera images for data collection due to practical and methodological reasons. First, the physical marker used for 6DoF pose ground truth is only reliably detected within limited viewing angles, and requiring visibility in both stereo views would constrain pose diversity. Second, applying image inpainting to both stereo images could introduce inconsistencies that affect stereo matching and depth estimation. Finally, using stereo data increases the risk of calibration errors between cameras, which could degrade data quality. Focusing on the left images avoids these issues and improves dataset robustness.

\subsection{Future prospects}
\textbf{Dataset Improvement:} SurgRIPE dataset focuses on instrument pose estimation for monocular endoscopic images. Considering da Vinci\textregistered endoscope can capture stereo images, more stereo images could be collected in the future to utilize extra depth information from stereo images. In addition, the dataset can be improved by adding more realistic tissue as background to simulate real surgery. So far, the target for the pose estimation is the wrist joint between the shaft and the end effector of the tool. The tool pose is estimated with respect to the camera coordinate system, which is our reference coordinate system. Since the end effector can only rotate along the z-axis of the wrist joint, the pose of the tool tip can be easily recovered using the estimated wrist joint pose by detecting the 2D position of the tool tip and applying simple 3D geometry.

\section{Conclusions}  \label{conclusions}
This paper presents the SurgRIPE challenge, which is part of the Endoscopic Vision Challenge, organised in conjunction with MICCAI2023. This work focuses on the markless 6DoF surgical instrument pose estimation with and without occlusion. We first introduce a 6DoF pose estimation dataset for surgical instrument pose estimation along with a benchmark framework to comprehensively evaluate pose estimation methods. Our validation framework is expected to be used as the standard benchmark framework for surgical instrument research.
\subsection{CRediT authorship contribution statement}
\textbf{Haozheng Xu}: Conceptualization, Methodology, Software, Resources, Data Curation, Formal analysis, Investigation, Validation, Visualization, Funding acquisition, Writing - Review \& Editing, Project administration.
\textbf{Alistair Weld}: Software, Formal analysis, Presentation, Investigation, Validation, Visualization, Writing - Review \& Editing, Project administration.
\textbf{Chi Xu}: Software, Validation, Presentation, Visualization.
\textbf{Alfie Roddan}: Software, Validation, Presentation, Visualization.
\textbf{Jo\~ao Cartucho}: 
Writing - Original Draft.
\textbf{Mert Asim Karaoglu}: 
Investigation, Writing - Original Draft.
\textbf{Alexander Ladikos}: 
Investigation, Writing - Original Draft.
\textbf{Yangke Li}: 
Investigation, Writing - Original Draft.
\textbf{Yiping Li}: 
Investigation, Writing - Original Draft.
\textbf{Daiyun Shen}: 
Investigation, Writing - Original Draft.
\textbf{Shoujie Yang}: 
Investigation, Writing - Original Draft.
\textbf{Geonhee Lee}: 
Investigation, Writing - Original Draft.
\textbf{Seyeon Park}: 
Investigation, Writing - Original Draft.
\textbf{Jongho Shin}: 
Investigation, Writing - Original Draft.
\textbf{Young-Gon Kim}:  
Investigation, Writing - Original Draft.
\textbf{Lucy Fothergill}: 
Investigation, Writing - Original Draft.
\textbf{Dominic Jones}: 
Investigation, Writing - Original Draft.
\textbf{Pietro Valdastri}: 
Investigation, Writing - Original Draft.
\textbf{Duygu Sarikaya}: 
Investigation, Writing - Original Draft.
\textbf{Stamatia Giannarou}: 
Conceptualization, Writing - Review \& Editing, Supervision.

\section*{Acknowledgments}
Dr Stamatia Giannarou, Mr Chi Xu and Mr Haozheng Xu are supported by the Royal Society [URF$\setminus$R$\setminus$201014]. Mr Alistair Weld is supported by the UK Research and Innovation (UKRI) Centre for Doctoral Training in AI for Healthcare (EP/S023283/1).

\section*{Compliance with ethical standards}
\textbf{Conflict of interest} The authors declare that they have no conflict of interest to disclose. \\
\textbf{Ethical approval} All human and animal studies have been approved and performed in accordance with ethical standards.\\
\textbf{Informed consent} All the data used for this publication was previously publicly available online, and was obtained with informed consent.\\
\textbf{Funding}
The authors are grateful for the sponsorship from Intuitive\textsuperscript{\tiny\textregistered} for the prize awards.

\bibliographystyle{model2-names.bst}\biboptions{authoryear}
\bibliography{references}

\end{document}